\documentclass[lettersize,journal]{IEEEtran}

\usepackage{amsmath,amsfonts}
\usepackage{algorithmic}
\usepackage{array}
\usepackage[caption=false,font=normalsize,labelfont=sf,textfont=sf]{subfig}
\usepackage{textcomp}
\usepackage{stfloats}
\usepackage{url}
\usepackage{verbatim}
\usepackage{graphicx}
\usepackage{cite}
\usepackage{amsmath,amssymb,amsfonts}
\usepackage{textcomp}
\usepackage[ruled,vlined]{algorithm2e}
\usepackage{tablefootnote}
\usepackage{arydshln}

\usepackage[english]{babel}
\usepackage{blindtext}
\begin{document}

\title{Efficient Robot Skill Learning with Imitation from a Single Video for Contact-Rich Fabric Manipulation}

\author{Shengzeng Huo, Anqing Duan, Lijun Han, Luyin Hu, Hesheng Wang and David Navarro-Alarcon
\thanks{This work is supported by the Research Grants Council (RGC) under grant 15212721. \textit{Corresponding author: D. Navarro-Alarcon}.}
\thanks{S. Huo, A. Duan and D. Navarro-Alarcon are with The Hong Kong Polytechnic University, Department of Mechanical Engineering, Hong Kong.}
\thanks{
L. Hu is with The University of Hong Kong, Centre for Transformative Garment Production, NT, Hong Kong.
}
\thanks{L. Han and H. Wang are with the Shanghai Jiaotong University, Department of Automation, Shanghai, China.}
}

\markboth{}
{HUO \MakeLowercase{\textit{et al.}}: Efficient Robot Skill Learning with Imitation from a Single Video for Contact-Rich Fabric Manipulation}


\maketitle


\begin{abstract}
Classical policy search algorithms for robotics typically require performing extensive explorations, which are time-consuming and expensive to implement with real physical platforms.
To facilitate the efficient learning of robot manipulation skills, in this work, we propose a new approach comprised of three modules: (1) learning of general prior knowledge with random explorations in simulation, including state representations, dynamic models, and the constrained action space of the task; (2) extraction of a state alignment-based reward function from a single demonstration video; (3) real-time optimization of the imitation policy under systematic safety constraints with sampling-based model predictive control. 
This solution results in an efficient one-shot imitation-from-video strategy that simplifies the learning and execution of robot skills in real applications. 
Specifically, we learn priors in a scene of a task family and then deploy the policy in a novel scene immediately following a single demonstration, preventing time-consuming and risky explorations in the environment. 
As we do not make a strong assumption of dynamic consistency between the scenes, learning priors can be conducted in simulation to avoid collecting data in real-world circumstances.
We evaluate the effectiveness of our approach in the context of contact-rich fabric manipulation, which is a common scenario in industrial and domestic tasks.
Detailed numerical simulations and real-world hardware experiments reveal that our method can achieve rapid skill acquisition for challenging manipulation tasks.

\end{abstract}


\section{Introduction}
\IEEEPARstart{R}{einforcement} learning (RL) algorithms are becoming more and more popular since they allow robots to learn by trial and error, avoiding troublesome modeling of specific situations \cite{rl_survey_tnnls}. 
However, it is difficult to directly transfer these RL methods to robots since these algorithms generally require millions of interactions \cite{Chatzilygeroudis2018ASO}.
Previous works have leveraged imitation learning to speed up the learning process, such as pouring \cite{one_shot_pour_tase}, nonprehensile manipulation \cite{il_tase_push}, and contact-rich manipulation \cite{il_tase_contact}.
Most of them require a large database comprising state-action tuples and additional explorations for policy learning. 
However, these two requirements are difficult to satisfy for a complex manipulation task in an unstructured environment since: 
(1) high-dimensional visuomotor policy is data-hungry \cite{Zhang2017DeepIL}; (2) random explorations in unstructured environments are dangerous for real robots \cite{Mitrano2021LearningWT}. 
An ideal imitation scenario is one which is similar to the way humans learn from others, e.g., a student observes and imitates a skill from a teacher \cite{dome} or even a video on the web \cite{youtube}. 
Achieving this level of sensorimotor dexterity is precisely our goal in this paper. We develop a one-shot observational imitation approach that enables robots to deploy a task immediately. 

\begin{figure}
\centering
\centerline{\includegraphics[width=\columnwidth]{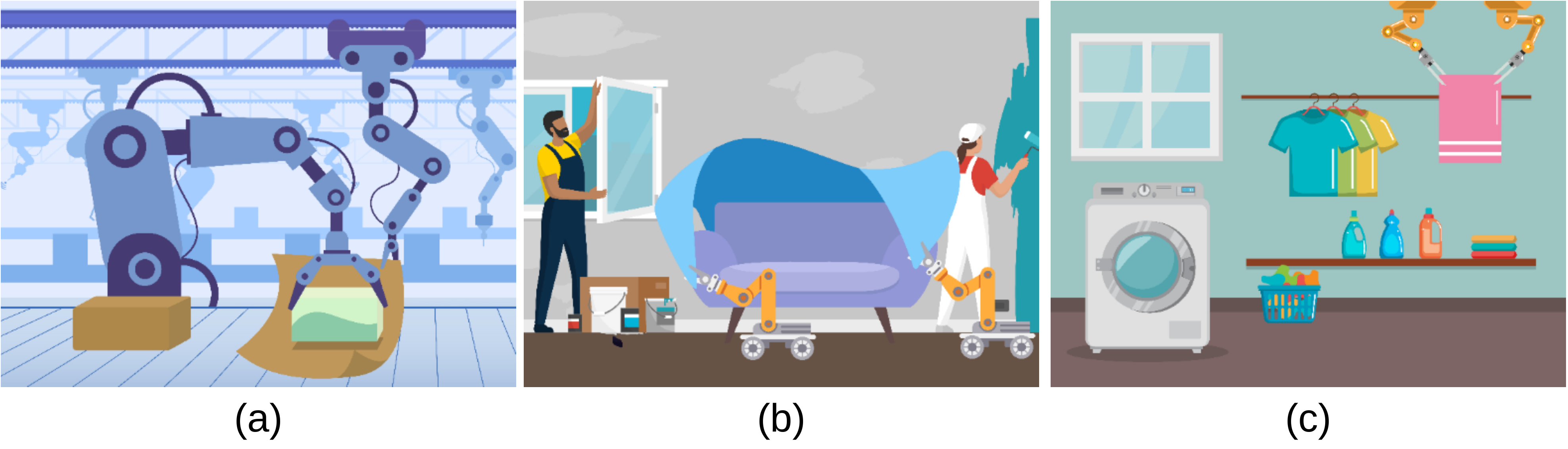}}
    \caption{Scenarios of contact-rich fabric manipulation. (a) Packing. (b) Covering. (c) Hanging.
    }
    \label{1_scene}
\end{figure}

One of the factors contributing to human beings' strong imitation ability is that we extract the temporal abstraction from demonstrations rather than precise state-action pairs \cite{state_align}.
Motivated by this understanding, our core idea of efficient robot skill learning is to follow the state sequences of a new scene in the demonstration as closely as possible. 
Two strategies are designed to achieve this goal: (1) learning general prior knowledge about the task family in a simulated scene; (2) taking long-term estimated resulting states under constraints into account with MPC. 
We emphasize that there are differences between sim-to-real RL methods \cite{matas2018sim} and our efficient skill-learning approach. 
The first one aims to create a simulation configuration that closely resembles the actual task before learning a policy for that specific activity.
Compared with them, our goal is to enable robots to deploy a new scene in the task family, which does not appear in the simulation. The benefits of our formulation are as follows: (1) our policy can generalize to different scenes within the task family efficiently (with only a demonstration video); (2) we do not demand a very realistic simulation environment and additional sim-to-real skills.

Our approach is validated in the context of contact-rich fabric manipulation, which means the deformable fabric needs to interact with rigid objects and establish symbolic object-object relationships with each other. 
Fig. \ref{1_scene} presents various industrial and household scenarios of this task, revealing that it is an essential skill for robots.
We choose this task for two main reasons. For one thing, it is difficult to solve with the conventional end-to-end imitation learning methods since (1) a great scale of demonstrations is required due to the high-dimensional state (infinite degrees of freedom of fabrics) and action (bimanual manipulation) space; and (2) random explorations should be prohibited due to the potential for collisions in contact-rich environments. For another, the superiority of our approach can be illustrated with this specification since (1) transferring between various interaction scenarios provides strong evidence of the model's inference ability; (2) the symbolic object-object association goal requires high-level planning instead of mathematical approximation.
The original contributions of this work are:
\begin{itemize}
    \item We formulate the efficient robot skill learning problem as the process of adapting from learned priors of the task family to a new scene.  
    \item We design a state alignment-based reward function to extract temporal abstraction from a demonstration video.
    \item We leverage a sampling-based MPC method to generate promising and safe motions in imitation.
    \item We report a detailed experimental study to evaluate the performance of the proposed approach.
\end{itemize}

In contrast with most algorithms in the literature, our approach endows robots to perform a novel scene within the task family effectively.
Specifically, the guidance of the new scene is only a single observational demonstration and robots can conduct it immediately without any explorations.
The proposed approach could advance the development of efficient robot skill learning.


The rest of this paper is organized as follows. Sec. II introduces the related work. Sec. III states the problem formulation. Following that, we explain the details of our methodology in Sec. IV. 
Sec. V reports the validation experiments and the corresponding analysis. Finally, Sec. VI concludes this article.

\section{Related Work}
\subsection{Imitation Learning from Observation}
Standard imitation learning methods, either behavior cloning (BC) or inverse reinforcement learning (IRL), assume both observations and actions are available in the demonstrations. In general, two methods are used to collect the proprioceptive information of the robots.
For one thing, operators directly contact and guide the robot through kinesthetic teaching, which is not suitable for robots with multiple degrees of freedom. 
For the other, robots are teleoperated by remote control devices, such as VR \cite{Zhang2017DeepIL} and mobile phones \cite{mandlekar2018roboturk}.
However, additional efforts of setting and training are required for the operator, affecting the practicality in real applications.


To simplify the demonstration procedure, imitation from observation (IfO) has garnered a great deal of attention, in which only the observation sequence is required. However, it brings additional challenges in imitation due to the lack of action information.
A natural solution is recognizing the movement in the video to learn the policy in a conventional manner (e.g. \cite{bc_from_obs, Bahl2022HumantoRobotII}).
Other researchers \cite{xirl, Mees2019AdversarialSN} establish auxiliary rewards for imitation through learning the representation and the correspondence of the observations.
The main drawback of the above methods is the requirement of many demonstrations to guarantee generalizability.
To learn a robot skill efficiently, more and more researchers have switched their focus to one-shot imitation learning.

\subsection{One-shot Imitation Learning}
One-shot imitation learning means robots are able to learn from a single demonstration of a given task and then generalize to new situations of the same task \cite{duan2017one}. However, it is very challenging due to data scarcity.
One solution is to incorporate imitation learning with RL to improve the robustness of the policy with fine-tuning.
\cite{inverse_dynamic_one_shot} integrates a task-specific inverse dynamic model into RL.
\cite{align_tool} extracts the tool trajectory from the video and aligns the simulated environment with the video to initialize a policy for RL to learn. There are two main drawbacks of this kind of method: (1) training robots in the real world is challenging due to sample efficiency and safety concerns. (2) learning in simulation necessitates a high level of verisimilitude with the actual world and additional techniques to deal with the sim-to-real gap.
Other researchers \cite{assemble_ifo_tii, Johns2021CoarsetoFineIL} formulate this problem as visual servoing. \cite{assemble_ifo_tii} iterates the pose aligning between the robot and the human hand in the video; \cite{Johns2021CoarsetoFineIL} obtains the approaching policy in a self-supervised manner through backward learning from the goal pose. However, the above methods hold strong task-related assumptions (e.g. known specific parameters \cite{assemble_ifo_tii}, and proximity to goal pose \cite{Johns2021CoarsetoFineIL}).

In addition, some researchers formulate one-shot imitation in a meta-learning manner \cite{finn2017model}, which means learning a general policy with few diverse demonstrations for different tasks, then adapting the policy for a new task given a single demonstration \cite{one_shot_pour_tase}. However, collecting such a big database is time-consuming and troublesome, especially in reality. Our work is built upon this formulation, in which we only learn prior knowledge in a single scene of the task family through random explorations and adapt to a novel scene according to a demonstration video. 


\subsection{Contact-Rich Fabric Manipulation}
Many researchers have addressed the robotic manipulation of soft materials, see \cite{yin2021modeling,survey_jihong} for recent reviews. However, most existing methods (either model-based \cite{Zhang2022LearningGM, Wang2022OfflineOnlineLO} or model-free \cite{8106734, Wang2018AdaptiveVS}) only consider shaping tasks without complex contact with objects in structured environments.
Contact-based manipulation of a deformable linear object towards a desired shape is considered in \cite{Huo2022KeypointBasedPB}, while customized action primitives are required for the controller.
Although fabric manipulation has gained interest from several researchers, most of them focused on laundry folding \cite{lippi_fold} and flattening \cite{ha2021flingbot}. 
\cite{mcconachie2020manipulating} estimates the model utility of fabrics in unstructured environments, while the manipulation strategy is only validated in simulation.
The method in \cite{Seita2018DeepTL} solves a bed-making task with strong assumptions about the initial configuration and customized placing policy.
\cite{matas2018sim} exploits model-free reinforcement learning to learn a policy of hanging a towel over a hanger. However, it requires a high degree of verisimilitude in the simulation environment and a great demand for exploration resources.

Due to the planning complexity of DOM, some researchers (e.g. \cite{Salhotra2022LearningDO, Seita2021LearningTR}) have leveraged imitation learning to improve the learning efficiency, while most of them require multiple demonstrations comprised of state-action transitions. Moreover, most of them only deal with simple tasks in a structured environment \cite{Salhotra2022LearningDO} or only implement the policy in simulation \cite{Seita2021LearningTR}. 
Hence, there are still open questions about how to acquire efficient robot skills for contact-rich fabric manipulation. 

\section{Problem Formulation}
\begin{figure*}
\centering
\centerline{\includegraphics[width = \textwidth]{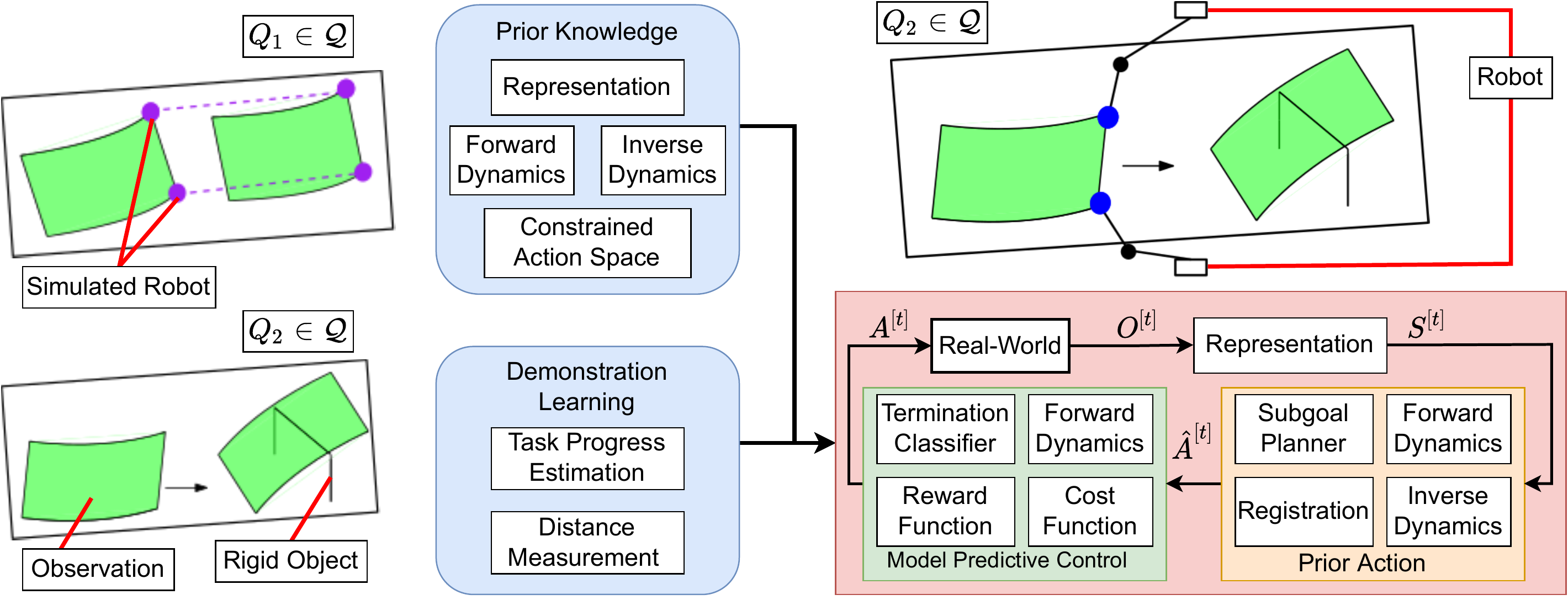}}
    \caption{
    The overview of our efficient robot skill learning approach. 
    With the prior knowledge learning in the scene $Q_1$ within the task family $Q_1\in\mathcal{Q}$, the model predictive control method generates motions for the novel scene $Q_2\in\mathcal{Q}$ guided by the reward function from the demonstration learning.
    }
    \label{2_framework}
\end{figure*}
For a robotic task in real-world, we formulate the problem as a Partially Observable Markov Decision Process (POMDP) represented by a tuple $\mathcal{M} = (\mathcal{S}, \mathcal{O}, \mathcal{A}, \mathcal{P}, R, \rho_0, \gamma)$, where the state $S^{[t]}\in\mathcal{S}$ is unknown and its corresponding observation is $O^{[t]}\in\mathcal{O}$. State transition function $\mathcal{P}(S^{[t+1]}|S^{[t]}, A^{[t]})$ characterizes the probability of a switch from state $S^{[t]}$ to $S^{[t]+1}$ after executing action $A^{[t]}\in\mathcal{A}$ and $R(S^{[t]}, A^{[t]})$ is the reward function.
$\rho_0$ is the initial state distribution and $\gamma$ is the discount factor. 

For the reward function, we consider the sparse terminal configuration, which is a common situation in robot tasks \cite{Wang2020LearningOL}. This means that robots need to perform correct actions continuously and finally complete the task to obtain a positive reward. The specific definition is:
\begin{equation}
R(S^{[t]},A^{[t]})= \begin{cases}1, & \text { if the task is completed } \\ 0, & \text { else }\end{cases}
    \label{reward}
\end{equation}
During the learning process, robots probably enter dangerous states that cause catastrophic results, especially for safety-critical applications \cite{Liu2020SafeMR}.
Hence, we induce a cost function to evaluate if an action leads to an unsafe state. Specifically, an action $A^{[t]}$ is valid if its resulting state $S^{[t+1]}$ is safe, denoted as:
\begin{equation}
    C(S^{[t]}, A^{[t]}) = \begin{cases}0, & \text { if } S^{[t+1]}\in\mathcal{S}_V \\ 1, & \text { else }\end{cases}
    \label{cost}
\end{equation}
where $\mathcal{S}_V\subseteq\mathcal{S}$ is the safe state subset. According to this definition, we constrain the feasible subset $\mathcal{A}_V^{[t]}$ within the entire action space $\mathcal{A}$ with respect to the current state $S^{[t]}$, denoted as $\mathcal{A}_V^{[t]}\subseteq\mathcal{A}$.

Constrained reinforcement learning is a general method for policy search under certain conditions \cite{Liu2020SafeMR}, which aims to maximize the cumulative reward while limiting the cost incurred from constraint violations:
\begin{equation}
    \pi^*=\arg\max_\pi [J_R(\pi) -  J_C(\pi)] 
\end{equation}
where $J_R(\pi)=\mathbb{E}_{\mathcal{T} \sim \pi}\left[\sum_{t=0}^T\gamma
^tR\left(S^{[t]},A^{[t]}\right)\right]$ and $J_C(\pi)=\mathbb{E}_{\mathcal{T} \sim \pi}\left[\sum_{t=0}^T C\left(S^{[t]},A^{[t]}\right)\right]$ denote the expected return of policy $\pi$ with respect to the reward function $R$ and the cost function respectively.

However, there are two main problems when applying RL in a real application: safe learning and sample efficiency. Human supervision and manual reset are necessary to avoid unacceptable catastrophes for robots, which are time-consuming and labor-costing for operators due to sample inefficiency. To deal with these issues, we formulate this problem as learning priors in a scene $Q_1$ of the task family $Q_1\in\mathcal{Q}$ and then deploy a new scene of the same family $Q_2\in\mathcal{Q}$ with a demonstration provided, as shown in Fig. \ref{2_framework}. We leverage the contact-rich fabric manipulation context to explain the idea of the formulation. We consider the interactions between a piece of fabric and different rigid objects as different scenes of the task family $Q\in\mathcal{Q}$. Specifically, $Q_1$ in Fig. \ref{2_framework} is a contact-free scene in simulation without rigid objects, while $Q_2$ requests the fabric to interact with a hanger to achieve the goal. The details about the policy transfer from a scene $Q_1$ to another scene $Q_2$ are delivered in Sec. IV.

Next, we introduce the specifications of the contact-rich fabric manipulation context, whose objective is to establish symbolic relationships between the deformable fabric and the rigid object. For example, the fabric covers the furniture entirely and stably as shown in Fig. \ref{1_scene}(b).
The following assumptions are made for the context:
\begin{itemize}
    \item The fabric is initially placed on a table with a flattened configuration and the rigid object is fixed on the table.
    \item The mask of the fabric in the scene is known.
    \item Dual arms rigidly grasp two ends of the fabric without any loose contact during manipulation.
    \item Dual arms are kinematically controlled and the coordination between them and the camera is calibrated.
\end{itemize}
We make the assumption about the fabric since there are a few existing works about unfolding \cite{ha2021flingbot} and the cloth region segmentation \cite{cloth_segmentation} is not our focus.
Other assumptions are commonly used in deformable object manipulation \cite{navarro2016automatic}.

Starting from a pre-grasped configuration, the manipulation action of our dual-arm robot $A=[\boldsymbol a_L^{\boldsymbol T} \ \boldsymbol a_R^{\boldsymbol T}]$ consists of the relative movements of individual end-effectors, denoted as $\boldsymbol{a}^{\boldsymbol T} = [\Delta x\ \Delta y\ \Delta z]$.
When the termination conditions are reached, two end-effectors finish the movement and open the grippers to release the fabric. 

In summary, our goal is to maximize the accumulated dense reward $R_E(S^{[t]}, \{S^{[j]}_E\}_{j=1}^T)$ w.r.t. a $H$-length action sequence $\mathcal{X}=(A^{[1]},\cdots, A^{[H]})$, while avoiding breaking out the constraints $C(S^{[t]}, A^{[t]})$, denoted as:
\begin{equation}
\begin{gathered}
\mathcal{X}^*=\arg \max _{A^{[1]}, \ldots, A^{[H]}} \mathbb{E}\left[\sum_{t=1}^H \gamma^t R_E\left( S^{[t]}, \{S^{[j]}_E\}_{j=1}^T\right)\right] \\
\text { s.t. } 
\quad S^{[t+1]}=\mathcal{P}\left(S^{[t]}, A^{[t]}\right)\ \quad \forall t \in\{1, \ldots, H\} \\ 
C\left(S^{[t]},A^{[t]}\right)=0
\quad \forall t \in\{1, \ldots, H\}
\end{gathered}
\label{optimization}
\end{equation}

\section{Methods}
\subsection{Overview}
Our objective is efficient robot skill learning for a new scenario, which eliminates the need for collecting a large amount of data in reality and additional explorations in the training phase following the demonstration.
Fig \ref{2_framework} illustrates the overview of our proposed approach, which can be divided into three phases: \textbf{Prior}, \textbf{Demonstration}, and \textbf{Control}. In the following, we first introduce the procedures of data collection for \textbf{Prior} knowledge in simulation without human participation. Then, we explain how to extract the high-level temporal abstraction of a novel scenario from a \textbf{Demonstration} video. Finally, we introduce the details of our sampling-based model predictive \textbf{Control} that is able to achieve an efficient and safe policy in real-time.


\subsection{Prior Knowledge}
\begin{figure}
\centering
\centerline{\includegraphics[width=0.95\columnwidth]{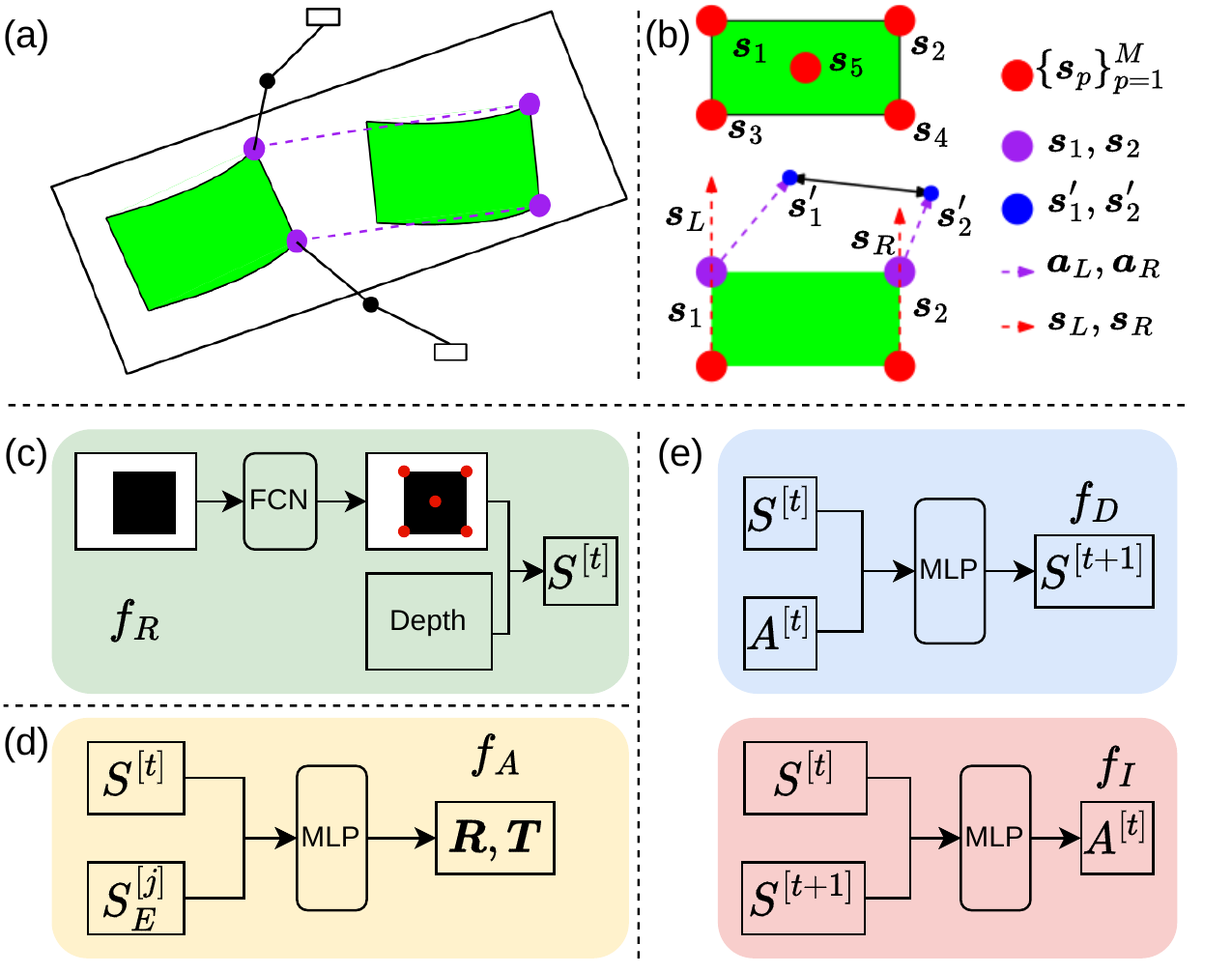}}
    \caption{Graphical illustration of the priors learning procedure. 
    (a) Bimanual manipulation in a contact-free scenario.
    (b) Details of the customized keypoints and the constrained action space. 
    The architecture of the neural networks. (c) State representation $f_R$. (d) State registration $f_A$. (e) Forward dynamics $f_D$ and inverse dynamics $f_I$. 
    }
    \label{3_priors}
\end{figure}
The objective of learning priors is to equip agents with key knowledge about a task family $\mathcal{Q}$, which is generalizable and suitable for any scene within the task family $Q\in\mathcal{Q}$. In this paper, three kinds of prior knowledge are considered: (1) state representation; (2) dynamic models, including forward and inverse dynamics; (3) definition of the constrained action space. 

To avoid specific modeling, we learn the descriptors and dynamics through neural networks in a data-driven manner. 
We collect the required data in simulation instead of reality since (1) collecting real-world data is time-consuming and costly; (2) simulation provides direct access to the state $S^{[t]}$ of the system, avoiding manual labeling. For our contact-rich fabric manipulation context, we learn priors in a contact-free simulation scenario as shown in Fig. \ref{3_priors}(a), in which two virtual anchors act as grippers in a planar configuration. The reasons for choosing this kind of configuration are (1) simulating the complex fabric-rigid object interactions with a high degree of verisimilitude is challenging to achieve; (2) interactions between fabric and various rigid objects share little similarity resulting in a poor generalization. The details about the simulation specification are introduced in Sec. V-A.
 

A mask of fabric in an RGB image plays as the raw observation $O^{[t]}$ in our context. 
Fig. \ref{3_priors}(b) shows our customized state description with $M$ keypoints $S^{[t]}=\{\boldsymbol s^{[t]}_p\}_{p=1}^M$, which is given directly in simulation. According to this semantic descriptor, we define the constrained action space $A^{[t]}\in\mathcal{A}^{[t]}_V$ with respect to an individual state $S^{[t]}$ for the entire contact-rich task family $\mathcal{Q}$. The basic principles of the constrained action space are: (1) ensuring the safety of the robots; (2) avoiding the severe self-occlusion of the fabrics during the manipulation.
In the following, we present the details of the constrained action space.

As shown in Fig. \ref{3_priors}(b), the positions of the end-effectors after executing the movement are denoted as $\boldsymbol s'_1 = \boldsymbol s_1 + \boldsymbol a_L$ and $\boldsymbol s'_2 = \boldsymbol s_2 + \boldsymbol a_R$, respectively.
Also, we define the edge vectors $\boldsymbol s_L = \boldsymbol s_1 - \boldsymbol s_3$ and $\boldsymbol s_R = \boldsymbol s_2 - \boldsymbol s_4$ corresponding to the actions $A=[\boldsymbol a_L^{\boldsymbol T} \ \boldsymbol a_R^{\boldsymbol T}]$, respectively.   
The rules of avoiding approaching dangerous states after executing an action $A^{[t]}$ are:
\begin{itemize}
    \item Two grippers keep an appropriate distance from each other, avoiding over-stretching or severe self-occlusion:
    \begin{equation}
       \tau_C\times L < ||\boldsymbol s'_1-\boldsymbol s'_2||_2 < \tau_F\times L
    \end{equation}
    where $L$ is the original length of the edge of the fabric, $\tau_C$ and $\tau_F$ are the ratio thresholds to measure the distance between two end-effectors with respect to $L$.
    \item The direction difference between the individual action $(\boldsymbol a_L, \boldsymbol a_R)$ and its corresponding edge of the fabric $(\boldsymbol s_L, \boldsymbol s_R)$ is constrained under the threshold $\tau_D$: 
    \begin{equation}
        \max(<\boldsymbol a_L,\boldsymbol s_L>, <\boldsymbol a_R,\boldsymbol s_R>) < \tau_D
    \end{equation}
    where $<\boldsymbol a, \boldsymbol b>$ computes the angle between two vectors and $\tau_D$ is an angle limitation threshold.
    \item The fabric locates within the workspace.
    \item Two grippers do not collide with the fixed rigid objects in the environment (if any).
\end{itemize}

To avoid troublesome human intervention, we sample random movements within the constrained action space $A^{[t]}\in\mathcal{A}^{[t]}_V$ at each time-step $t$.
Through the random explorations, we acquire several transitions storing in a dataset $\mathcal{D}=(O^{[t]}, S^{[t]}, A^{[t]}, O^{[t+1]}, S^{[t+1]})$, where $O^{[t]}$ is the masked image of the fabric and $S^{[t]}$ is the corresponding customized keypoints. 
Based on the collected dataset $\mathcal{D}$, we train our data-driven models. Firstly, we learn the representation $f_R(S^{[t]}|O^{[t]})$ with a Fully Convolution Network \cite{Huo2022KeypointBasedPB} (as shown in Fig. \ref{3_priors}(c)), which outputs the pixel coordinates of the keypoints $S^{[t]}$ in the mask image. Then, we obtain the 3D representation of the keypoints by querying the corresponding depth value. Finally, two fully-connected networks are established to approximate the forward $f_D(S^{[t+1]}|S^{[t]}, A^{[t]})$ and the inverse dynamics $f_I(A^{[t]}|S^{[t]}, S^{[t+1]})$ of the system on the basis of the predicted keypoints $S^{[t]}$, as shown in Fig. \ref{3_priors}(e). All the models are trained with the mean square error (MSE).

\subsection{Demonstration Learning}
\begin{figure}
\centering
\centerline{\includegraphics[width=\columnwidth]{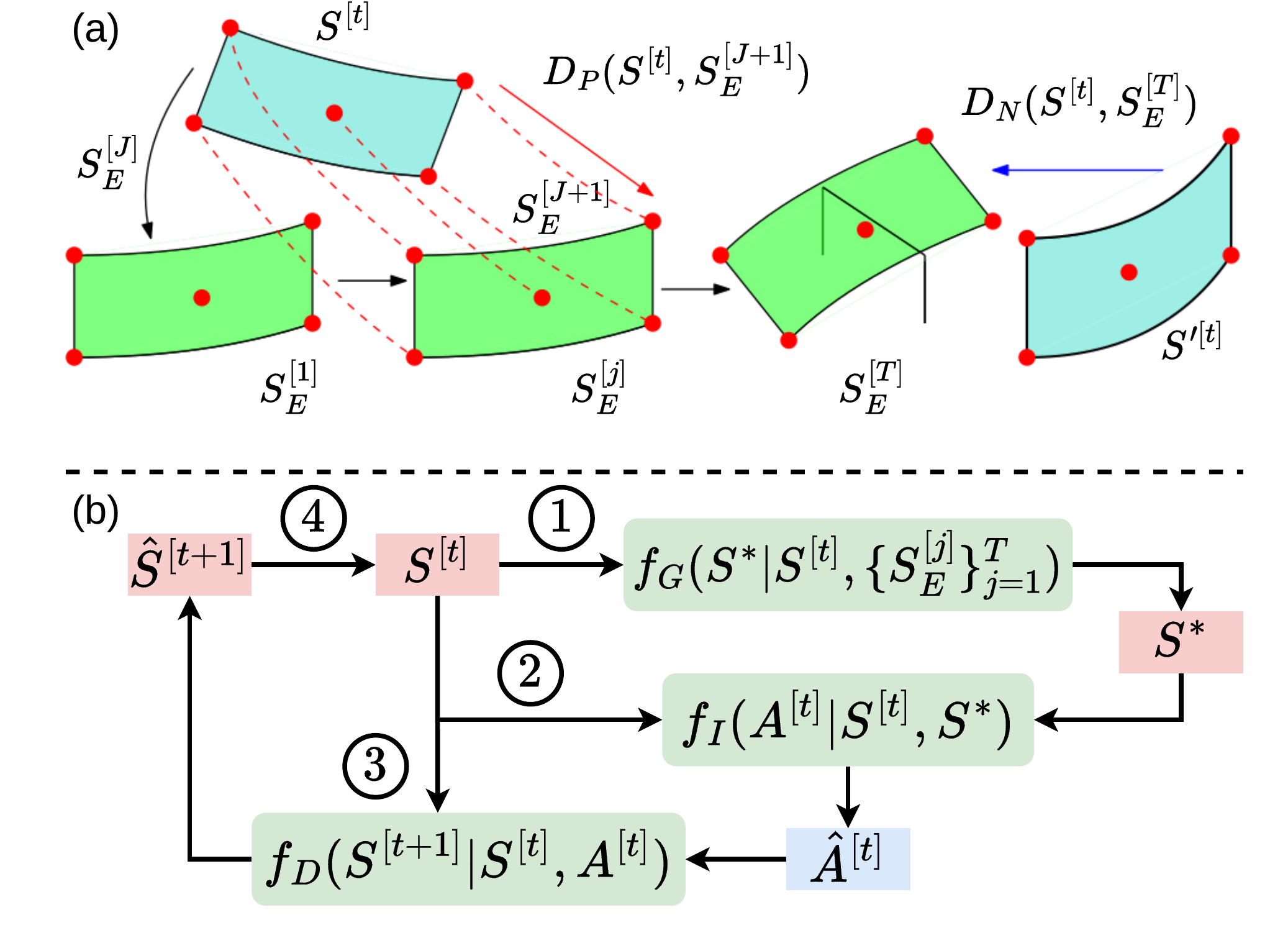}}
    \caption{(a) Graphical explanation of the state alignment-based reward function. The reward of a state $S^{[t]}$ is measured through registration with the demonstration. The state $S'^{[t]}$ that passes the final instance in the demonstration $S_E^{[T]}$ is penalized in the reward function. (b) The paradigm of the acquisition of the prior action.
    }
    \label{4_state_align}
\end{figure}
Instead of learning an end-to-end imitation policy directly, we emphasize the resulting state of each action 
since (1) learning the state-action mapping is data hungry; (2) the simulation-to-reality gap affects the performance of the end-to-end policy in transfer;
(3) the suitable action with respect to a state to approach the ultimate goal is generally not unique, especially for a high-dimensional task.
As a result, we convert the conventional end-to-end policy as searching for an optimal action whose forecast resulting state is closest to the state sequence in the demonstration.
\cite{Liu2019StateAI} holds a similar idea to us, while they are interested in an imitation scenario whose dynamics mismatch with the expert.
Compared with them \cite{Liu2019StateAI}, we consider a more complex scenario (inference to a novel scene of the task family), and the imitation is conducted immediately following a single demonstration video without any additional explorations.

Given the $T$-length single observation-only demonstration $\tau_E=\{O_E^{[j]}\}_{j=1}^{T}$, we first obtain the corresponding states $\{S_E^{[j]}\}_{j=1}^{T}$ with our representation model $f_R(S^{[t]}|O^{[t]})$.
Since the goal of the contact-rich task is to establish the symbolic relationship between the fabric and the rigid object, we set up a coordinate frame associated with the pose of the rigid object and represent all the states in this frame. The advantage of this setting is that the imitation is platform-agnostic, regardless of the configuration of sensors or actuators. 
Since the 6-DOF pose recognition of a rigid object is not a focus of our work, it is achieved through detecting a marker in real-world experiments.


Fig. \ref{4_state_align}(a) explains the state alignment process. 
Specifically, there are two critical issues to determine in the reward function: (1) estimate the temporal progress in the imitation; (2) measure the spatial distance to the final goal. 

For the first issue, we leverage the Intersection over Union (IoU) index to measure the similarity $D_E(S^{[t]}, S^j_E)$ between the current state $S^{[t]}$ and the states in the demonstration $\{S_E^{[j]}\}_{j=1}^T$.
\begin{equation}
    D_E(S^{[t]},S^j_E) = \frac{S^{[t]}\cap S^j_E}{S^{[t]}\cup S^j_E}
\end{equation}
However, there is no overlapped area between the state $S^{[t]}$ and the state sequence in the demonstration in some cases, which always happen at the early stage of the imitation. 
As a result, the estimator fails to distinguish between various states. 
Under this circumstance, we switch the similarity metric $D_E(S^{[t]}, S^j_E)$ to the Euclidean distance of the centers. 
According to the above metrics, the temporal progress estimation is implemented with:
\begin{equation}
    J = \arg\min_j D_E(S^{[t]},S^j_E)
    \label{task progress}
\end{equation}

We assume the provided demonstration is optimal; thus the imitator should follow the demonstration step-by-step. 
For this sequential imitation, we take two aspects into consideration for the second issue (as shown in Fig. \ref{4_state_align}(a), whose details are:
(1) \textbf{Positive}: the distance to the subgoal $S_E^{[J+1]}$, which is the next state of our temporal progress estimation result in the demonstration.
(2) \textbf{Negative}: the distance to the final state $S_E^{[T]}$ in the demonstration when the current state passes the last state in the demonstration $S_E^{[T]}$.
The first term encourages the agents to approach the next stage in the demonstration, while the second term penalizes the actions when the resulting state crosses the ultimate state in the demonstration. 
This crossing phenomenon ($S'^{[t]}$ in Fig. \ref{4_state_align}(b)) should be avoided since the manipulation in the opposite direction is required to return the state back to the common situation, which is not allowed in the constrained action space.
 
We implement registration between the current state $S^{[t]}$ and the subgoal $S_E^{[J+1]}$ to quantify their \textbf{Positive} similarity. 
A neural network $f_A (\boldsymbol R, \boldsymbol T|S^{[t]}, S^{J+1}_E)$ is utilized to output their relative transformation, including the rotation $\boldsymbol R$ and the translation $\boldsymbol T$. The detailed architecture is shown in Fig. \ref{3_priors}(d).
This registration model is trained by taking the states in the prior dataset $\mathcal{D}$ as the source and the demonstration $\{S_E^{[j]}\}_{j=1}^T$ as the target, respectively, whose goal is to minimize the MSE between $\{S_E^{[j]}\}_{j=1}^T$ and the transformed source $\boldsymbol R\cdot S^{[t]} +\boldsymbol T$:
\begin{equation}
    \mathcal{L}_A = ||\boldsymbol R\cdot S^{[t]} + \boldsymbol T - S^{[j]}_E||
\end{equation}
Based on the trained model, the distance scalar $D_P(S^{[t]},S_E^{[J+1]})$ is obtained through integrating the rotation $\boldsymbol R$ and the translation $\boldsymbol T$: 
\begin{equation}
\begin{aligned}
    D_P(S^{[t]},S_E^{[J+1]}) = |\boldsymbol T| + w_R\cdot|\boldsymbol R|
\end{aligned}
\end{equation}
where $\boldsymbol R, \boldsymbol T =f_A (\boldsymbol R, \boldsymbol T|S^{[t]}, S^{[J+1]}_E)$ and $w_R$ is the weight ratio to balance the rotation and the translation. Here, the rotation is represented with the Euler angle to correspond to the translation with different axes.

Compared with \textbf{positive} distance $D_P(S^{[t]}, S_E^{[J+1]})$, the \textbf{negative} distance $D_N(S^{[t]}, S_E^{[T]})$ is triggered only when the current state passes the last state in the demonstration, whose definition is: 
\begin{equation}
    D_N=
    \left\{\begin{array}{cc}
||S^{[t]}-S_E^{[T]}|| & J>T-2 \ \land\\
& ||S^{[t]}-S_E^{[T]}|| < ||S^{[t]}-S_E^{[T-1]}||  \\
0 & \text { else }
\end{array}\right.
\label{negative distance}
\end{equation}

Taking the above elements into consideration, the state alignment-based reward is denoted as:
\begin{equation}
\begin{aligned}
        R_E\left( S^{[t]}, \{S^{[j]}_E\}_{j=1}^T\right) = J - & w_P\cdot D_P(S^{[t]},S_E^{[J+1]}) \\
        & - w_N\cdot D_N(S^{[t]},S_E^{[T]})
\end{aligned}
\label{reward function}
\end{equation}
where $w_P$ and $w_N$ are the weight ratio of the positive distance and the negative distance, respectively.

\subsection{Model Predictive Control}
\begin{algorithm}
    \caption{State Alignment-driven MPC}
    \label{MPC}
    \textbf{Input:} Prior models $(f_R,f_D,f_I)$, Noise distribution $\epsilon\sim\mathcal{N}(\mu,\Sigma)$\\
    \textbf{Output:} Solution $A^{[t]}$ with the highest reward\\
    Collect the observation-only demonstration $\{O_E^{[j]}\}_{j=1}^T$ \\
    Obtain the state alignment-based reward function $R(S^{[t]},S_E) \leftarrow$ Eq. \ref{reward function} \\    
        	
    \While{is not terminated}
    {
    Get the action prior $\hat{\mathcal{X}}^{[t]}\leftarrow $ Eq. \ref{action prior} \\
    \While{the convergence is not met}
    {
        Sample $N$ solutions  $\Omega\in\{\mathcal{X}_i\}_{i=1}^N$ \\
        Evaluate the accumulated reward $\mathcal{R}(\mathcal{X}; S^{[t]})$ and the cost $\mathcal{C}(\mathcal{X}; S^{[t]})$ for each solution.\\
        Select and sort the feasible set $\Omega_V\subseteq \Omega$. \\
        Update the parameters $(\mu,\Sigma)\leftarrow$ Eq. \ref{update}
    }
    }
\end{algorithm}
Incorporating the prior knowledge from $Q_1\in\mathcal{Q}$, we hope that our robots deploy a novel task $Q_2\in\mathcal{Q}$ immediately after a demonstration video is given without any explorations. To this end, we take the safety issues of real robots into consideration and formulate the policy search as a constrained optimization problem. MPC is an effective method to deal with this issue. For example, \cite{7362013} considers the robotic visual servoing system's input and output constraints, while it depends on an analytical expression of the task.
\cite{Liu2020SafeMR} studies the safe RL problem with sparse indicator signals for constraint violations. However, it requires a few violation budgets to explore the environment and only deals with a task with original dense rewards in simulation. 
In this paper, we extend this method for efficient robot skill learning in reality in three aspects:
(1) customize a proper reward function and a cost function for safe and robust imitation from observation; 
(2) acquire a prior action to improve the efficiency of the sampling-based controllers;
(3) design a termination classifier that predicts if the imitation has been completed.

Taking the future expectation into consideration, MPC strives to obtain an $h$-length action sequences $\mathcal{X}^{[t]}=\{A^{[t+k]}\}^{h-1}_{k=0}$ through solving an open-loop optimal control problem.
To enable real-time control, it is important to have a high computation efficiency for each optimization step.
The sampling-based MPC approach in \cite{Liu2020SafeMR} searches within the entire action space $\mathcal{A}$, resulting in the low efficiency of the iteration scheme for a high-dimensional control policy.
To resolve this problem, we obtain a prior action sequence $\hat{\mathcal{X}}$ acting as a baseline to narrow down the search space.
The procedure of this paradigm is shown in Fig. \ref{4_state_align}(b).
For the current state $S^{[t]}$, the subgoal planner $f_G(S^*|S^{[t]},\{S_E^{[j]}\}_{j=1}^T)$ outputs a target $S^*$, which is actually $S_E^{[J+1]}$ computed with our temporal progress estimation model in Eq. \eqref{task progress}. Then we acquire an estimated action $\hat A^{[t]}$ at the current timestep $t$ with the inverse dynamic model $f_I$. 
\begin{equation}
    \hat{A}_{t} = f_I(S^{[t]},S^*), S^*=S_E^{J+1}
    \label{action prior}
\end{equation}
Since this imagined action will not be executed actually, we predict the achieving state $\hat S^{[t+1]}$ with the forward dynamic model $f_D$. 
Finally, this estimated state $\hat S^{[t+1]}$ will act as the new $S^{[t]}$ to start another new loop. The above procedure iterates $h$ steps to acquire a complete action sequence baseline $\hat{\mathcal{X}} = \{\hat{A}^{[t+k]}\}_{k=0}^{h-1}$.

Based on the prior action sequence $\hat{\mathcal{X}}$, each candidate of the sampling set $\mathcal{X}\in\Omega$ for the constrained optimization is generated through integrating the noise with it $\mathcal{X} = \hat{\mathcal{X}} + \epsilon$. In particular, $\epsilon$ is sampled from a $n$-dimensional factorized multivariate Gaussian distribution $\epsilon\sim\mathcal{N}$
($\mu,\Sigma$), where $\mu$ is the mean vector and $\Sigma$ is a diagonal covariance matrix. 

Next, the accumulated performance of each solution $\mathcal{X}\in\Omega$ is evaluated with our customized reward function $R_E\left( S^{[t]}, \{S^{[j]}_E\}_{j=1}^T\right)$ and the cost function $C(S^{[t]}, A^{[t]})$:
\begin{equation}
\begin{aligned}
&\mathcal{R}\left(\mathcal{X} ; S^{[t]}\right)=\sum_{k=1}^h \gamma^{k-1} R_E\left( S^{[t+k]}, \{S^{[j]}_E\}_{j=1}^T\right) \\
& \mathcal{C}\left(\mathcal{X}; S^{[t]}\right)=\sum_{k=1}^h   C\left(S^{[t+k-1]},A^{[t+k-1]}\right)
\end{aligned}
\label{update}
\end{equation}
where $S^{[t+k]}=f_D(S^{[t+k-1]},A^{[t+k-1]}),\forall k\in\{1,\cdots,h\}$ is predicted by the dynamic model $f_D(S^{[t+1]}|S^{[t]},A^{[t]})$. Then, we select the feasible solutions of all the candidates $\Omega_V\subseteq\Omega$ whose accumulated cost is zero. We sort these feasible solutions $\Omega_V$ and select the top samples to update the noise distribution $\epsilon\sim\mathcal{N}$
($\mu,\Sigma$) for the next iteration $i+1$ as:
\begin{equation}
\begin{aligned}
    \mu^{[i+1]} &\leftarrow (1-\beta)\mu^{[i]} + \beta \mu_V \\
    \Sigma^{[i+1]} &\leftarrow (1-\beta)\Sigma^{[i]} + \beta \Sigma_V 
\end{aligned}
\end{equation}
where $\beta$ is a hyper-parameter to determine the update portion for each iteration, $\mu_V$ and $\Sigma_V$ are the mean and the covariance of the noise corresponding to the feasible set $\Omega_V$. 
The iteration scheme is stopped when the optimal solution is feasible $C\left(\mathcal{X}; S^{[t]}\right)=0$ and the variance of the noise $\epsilon$ reaches the convergence: 
\begin{equation}
    \max\Sigma<\tau_C
\end{equation}

As a receding horizon control method, agents only apply the first input in the solution while others are discarded.

The outline of the control law is presented in Alg. \ref{MPC}. 
After observing a single demonstration video, dual arms execute the generated motion from the controller at each time-step $t$ to approach the ultimate goal continuously. An episode is terminated and the grippers are released if one of the situations occurs: (1) no feasible action under cost constraints is found; (2) exceeding the maximum exploration steps; (3) the goal is considered as finished from the terminal classifier. 
Since the goal specifications are diverse in the task family and challenge to model accurately in some complex instances,
we instead exploit the similarity between the achieved state $S^{[t+1]}$ and the final state in the demonstration $S^{[T]}_E$ to classify the goal completion.
Specifically, the task is considered finished when two conditions are satisfied simultaneously: (1) the IoU between $S^{[t+1]}$ and $S^{[T]}_E$ is larger than the pre-defined threshold $\tau_I$; (2) the temporal progress estimation (Eq. \eqref{task progress}) of $S^{[t+1]}$ matches with $S^{[T]}_E$.

\section{Results}
To assess our strategy comprehensively, we undertake statistical comparisons of our approach versus baselines and ablations in simulation. We begin by outlining the details of the simulation and a comparison study with several baselines. Then, an ablation study is implemented to evaluate the necessity of each component in our approach.
Finally, we demonstrate the practicality of our method for efficient robot skill learning in real applications.

\subsection{Simulation Setting}
\begin{figure}
\centering
\centerline{\includegraphics[width=0.9\columnwidth]{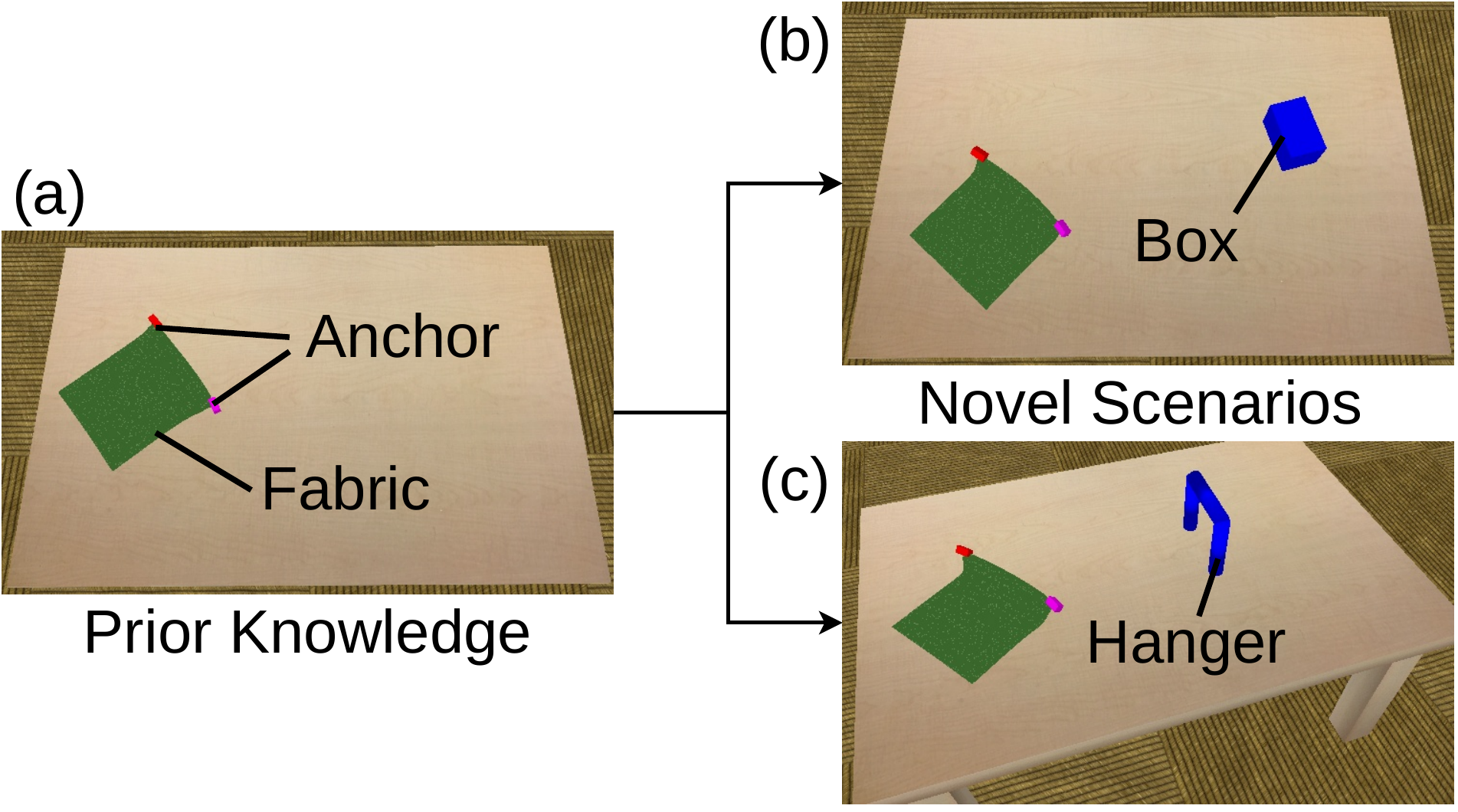}}
    \caption{Screenshots of the simulation environment. (a) Contact-free configuration. (b) \textbf{Box}: Cover a box. (c) \textbf{Hang}: Hang up a Fabric.
    }
    \label{5_simulation}
\end{figure}

\begin{figure}
\centering
\centerline{\includegraphics[width=\columnwidth]{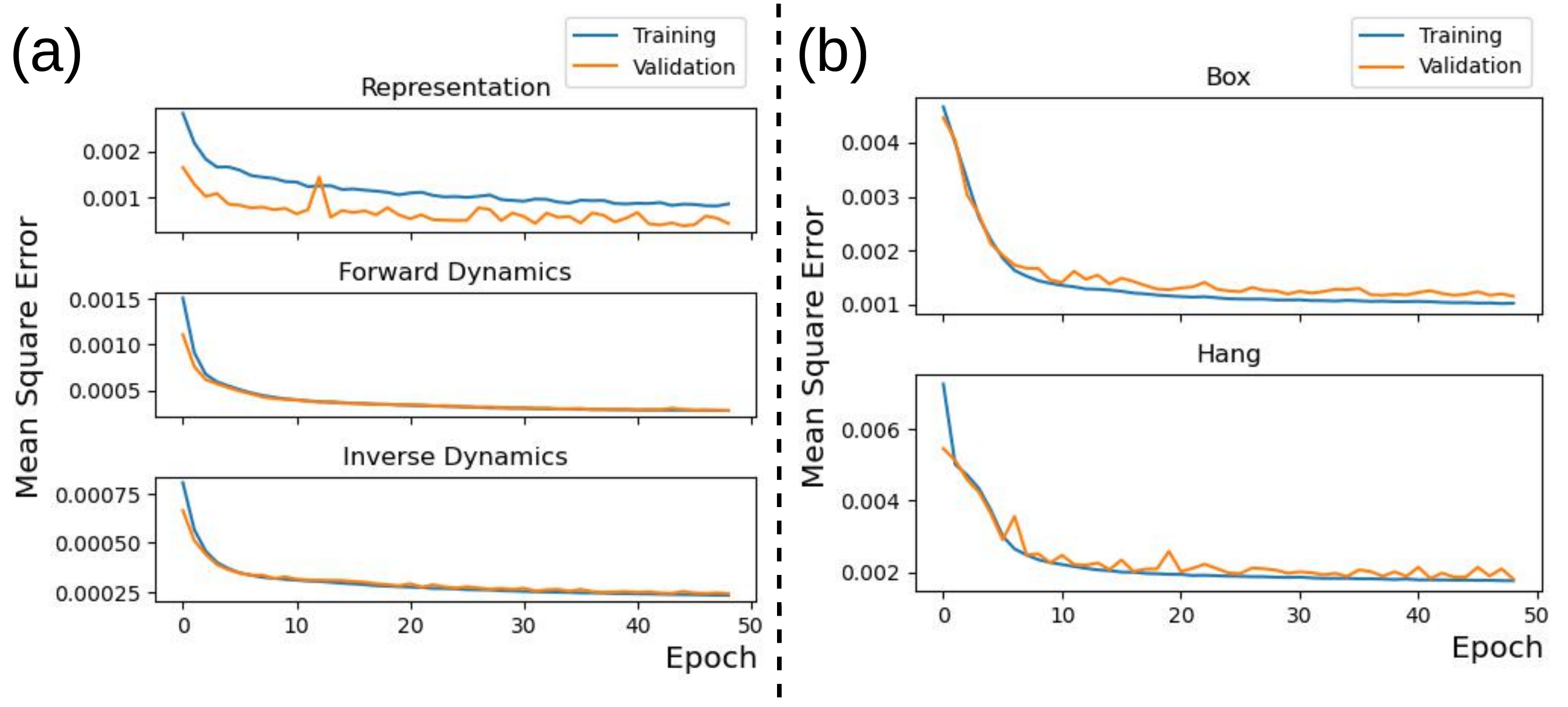}}
    \caption{The training and validation errors for the prior models. (a) Representation, forward and inverse dynamics. (b) Registration.
    }
    \label{6_training_curve}
\end{figure}

Fig. \ref{5_simulation} illustrates our simulation environment based on DeDo \cite{Antonova2021DynamicEW}, in which fabric is represented as a deformable grid mesh with 256 vertices. Two anchors attached to the corners of the fabric rigidly are used to render the grasping of two end-effectors.

To train the prior models in Sec. IV-B, we generate $10000$ transitions $(O^{[t]},S^{[t]},A^{[t]},O^{[t+1]},S^{[t+1]})$ through sampling random actions within the valid space $\mathcal{A}_v$ at each time-step $t$, as shown in Fig. \ref{5_simulation}(a). Note that there are no rigid objects to interact with the fabric in this environment. 


Two common contact-rich fabric manipulation scenarios, namely covering a box (\textbf{Box}) and hanging a towel (\textbf{Hang}), are established in simulation to evaluate the performance of the methods, as shown in Fig. \ref{5_simulation}(b) and Fig. \ref{5_simulation}(c), respectively. Note that these two environments are used to evaluate the algorithm instead of learning the specific models to transfer to the same task in reality.
In the simulation, we provide the demonstration by controlling the anchors while the action information is not recorded.
Rigid objects in both environments are fixed on the table. 
Each task is considered as successful if the symbolic relationship between the fabric and the rigid object is established after releasing the grippers. 
Starting from a pre-grasp configuration, the virtual grippers move and release at the end of an episode.

\subsection{Comparisons with Baselines}

\begin{figure*}
\centering
\centerline{\includegraphics[width=\textwidth]{"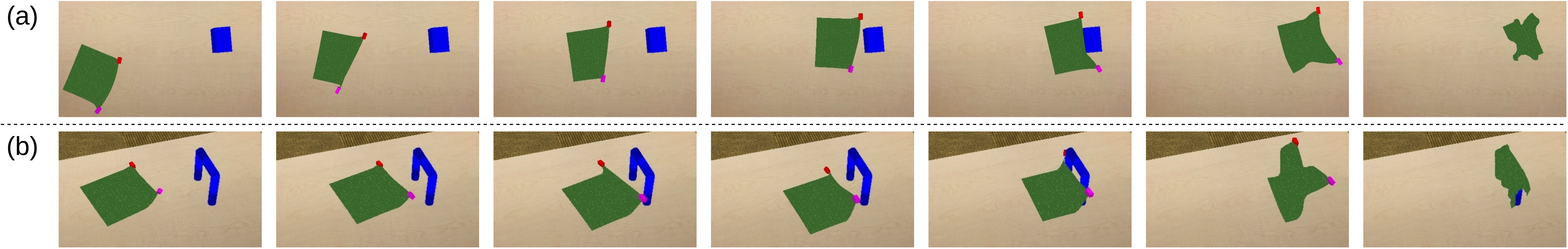"}}
    \caption{
    The scenarios of the contact-rich fabric manipulation tasks in simulation. (a) \textbf{Box}. (b) \textbf{Hang}.
    }
    \label{7_scene_simulation}
\end{figure*}
\begin{figure}
\centering
\centerline{\includegraphics[width=\columnwidth]{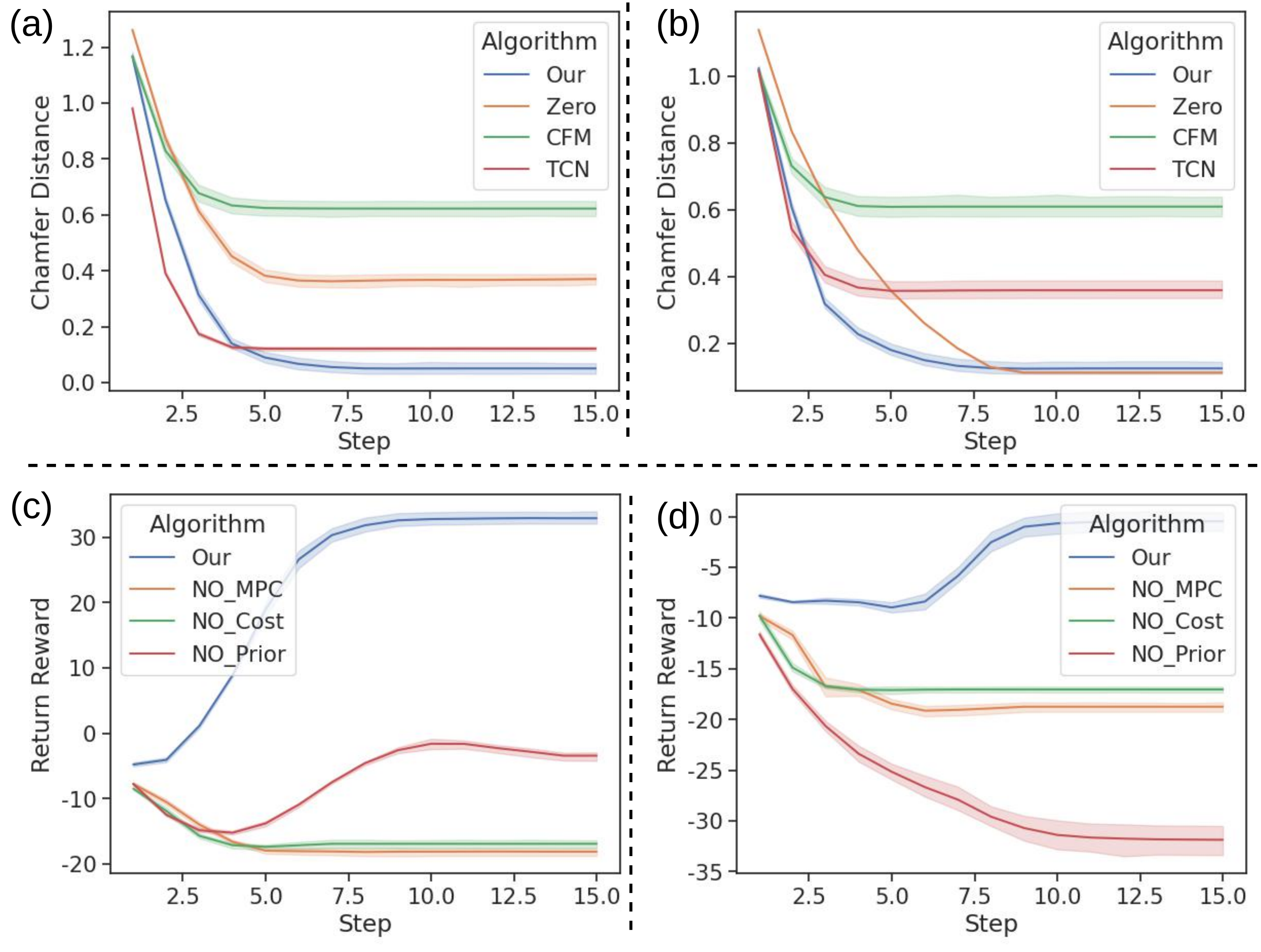}}
    \caption{The Chamfer distance comparison of several baselines. (a) \textbf{Box}. (b) \textbf{Hang}. The accumulated return comparison in the ablation study. (c) \textbf{Box}. (d) \textbf{Hang}.
    }
    \label{8_comparison_curve}
\end{figure}
In this section, we first present the achieved results of our approach. Then, we compare our approaches with other baselines.

The training and validation loss for the prior models (state representation and dynamic models) and the registration model for demonstration learning are visualized in Fig. \ref{6_training_curve}(a) and (b) respectively. All of the training only requires $20$ epochs.
To analyze our approach in detail, we provide two typical successful examples of individual contact-rich tasks in the trials, visualized in Fig. \ref{7_scene_simulation}(a) and (b) respectively. 
With a random initial configuration, the agents strive to align the states of the fabric to the state sequences in the demonstration. 
When the fabric is close to the rigid object, two virtual end-effectors follow the state sequences to lift up the object to avoid obstacle collision.
The attachment interactions disappear when the termination classifier returns a positive value and the fabric falls down slowly due to its gravity.

The goal of our approach is efficient robot skill learning without access to the actual scenario in the training phase. Specifically, robots learn the control policy from an offline simulated dataset instead of explorations in the exact environment. Moreover, robots should distill the learned knowledge in a scene of the task family and adapt to a new scene guided by a single observational demonstration. To this end, our approach incorporates prior knowledge and a state-alignment reward function into the sampling-based MPC method to obtain a safe and robust control policy.
To show the substantial improvements in our methods corresponding to the above arguments, we compare our method against various baselines, including \textbf{Zero} \cite{Pathak2018ZeroShotVI}, \textbf{CFM} \cite{yan2021learning} and \textbf{TCN} \cite{Sermanet2017TimeContrastiveNS}. The details of individual baselines are:
\begin{itemize}
    \item \textbf{Zero} learns a goal-conditioned skill policy with a forward consistency loss and then mimics the expert from a sequence of demonstration images.  
    \item \textbf{CFM} optimizes the visual representation and the dynamics with contrastive learning and then implements a model-based one-step optimal predictive control.
    \item \textbf{TCN} disambiguates temporal changes in the demonstration videos to provide a reward function for agents' policy search.
\end{itemize}
For a fair comparison, all the data-driven baselines share the same collected dataset $\mathcal{D}$ in simulation with respect to the task $Q_1\in\mathcal{Q}$ and a demonstration video for a novel task $Q_2\in\mathcal{Q}$. In addition, all the models are multi-layer perceptions (MLP) with two hidden layers of size $256$ followed by ReLU activation functions. 

Multiple trials are conducted to thoroughly assess the performance of the baselines. To evaluate the contact-rich fabric manipulation task performance quantitatively, we select Chamfer distance error \cite{Fan2016APS} as a metric to compare the achieved observation $O^{[t]}$ and the ultimate observation in the expert demonstration $O_E^{[T]}$, whose definition is: 
\begin{equation}
    d_{\mathrm{CD}}=\frac{1}{O_1} \sum_{x \in O_1} \min _{y \in O_2}\|x-y\|_2^2+\frac{1}{O_2} \sum_{y \in O_2} \min _{x \in O_1}\|y-x\|_2^2
\end{equation}
where $O_1$ and $O_2$ are two downsampled point clouds. 
Specifically, we acquire the 3D point cloud of the fabric by querying the corresponding depth value of the mask in the visual observation and down-sampling them to $N=200$ points equally.

Fig. \ref{8_comparison_curve}(a) displays the Chamfer Distance curve for different baselines and Table \ref{table_simulation} displays their success rates. Our method outperforms other baselines with respect to two individual tasks in success rate, while is comparable to \textbf{TCN} and \textbf{Zero} in terms of Chamfer distance in two tasks respectively. In the following, we present an analysis of this result.
With an embedding model to distinguish different states in the space, the spatial and temporal connection between them is not considered in \textbf{CFM}, resulting in poor performance in this complex sequential manipulation task.
We consider the failure of \textbf{TCN} is caused by the scarcity of data in a single demonstration, which hinders it to encode the temporal distance between the states and goal accurately.
Without a long-term prediction, it is easy for \textbf{Zero} to trap in a local distance minimum to the ultimate goal.
In addition, the comparable performance in terms of Chamfer Distance is mainly due to the constraint awareness in our approach. 
Based on the state alignment reward function, our approach seeks to reduce the error between the achieved results and the state sequences in the demonstration video step-by-step in a safe way under the cost function. 
However, the above baselines generate motions to go straight to the final goal. Although they can achieve a relatively small distance from the target, the desired symbolic object-object relationship is not established, resulting in low success rates. In other words, Chamfer Distance is only an auxiliary value to indicate the imitation process. This is also the reason why we need a state alignment-based reward to guide our control policy for this symbolic task.

\begin{table}
\centering
\caption{Comparisons between different methods}
\begin{center}
\begin{tabular}{|c|ccc|ccc|} 
\hline
\textbf{Method} & \multicolumn{3}{c|}{box} & \multicolumn{3}{c|}{hanger}  \\ 
 & Rate \% & $\mu_{IoU}$ & $\sigma_{IoU}$ & Rate \%  & $\mu_{IoU}$ & $\sigma_{IoU}$  \\ 
\hline
\textbf{Zero} & 1.8 & 0.02 & 0.01 & 0.0 &  0.00 & 0.00  \\
\textbf{CfM}  & 0.0 & 0.00 & 0.01 & 1.4 &  0.02 & 0.12  \\
\textbf{TCN}  & 6.6 & 0.08 & 0.22 & 2.4 &  0.03 & 0.14  \\
\hdashline
\textbf{No MPC}    & 0.0  & 0.00 & 0.01 & 3.0  & 0.03 & 0.17  \\ 
\textbf{No Prior}  & 14.4 & 0.15 & 0.33 & 7.4  & 0.07 & 0.24 \\ 
\textbf{No Cost}   & 3.0  & 0.03 & 0.16 & 2.2  & 0.02 & 0.14  \\ 
\textbf{Our}       & \textbf{81.9} & 0.74 & 0.38 & \textbf{72.7} & 0.69 & 0.42  \\
\hline
\end{tabular}
\end{center}
$\mu_{iou}$ and $\sigma_{iou}$ are the mean and the variance of IoU, respectively.
\label{table_simulation}
\end{table}

\subsection{Ablation Study}


In order to highlight the necessity of each component in our proposed approach, we implement a detailed ablation study. We contrast three alternative methods, including 
(1) \textbf{No MPC}: Adopt the output from the inverse dynamic model.
(2) \textbf{No Prior}: The MPC law iterates from scratch; 
(3) \textbf{No Cost}: The cost constraint function is removed in the MPC. 

According to the POMDP formulation,
we evaluate the performance of several ablations with the accumulated reward in this comparison. Note that agents will obtain a positive reward for episodic success and a negative warning reward due to the violation of the cost constraints.
We compare the accumulated rewards of alternative ablations in Fig. \ref{8_comparison_curve}(c)-(d) and compare their success rate of them in Table. \ref{table_simulation}. These results show that our suggested MPC approaches get the best success rate among them, while others are affected severely.

In the following, we provide a detailed analysis of their performance.
Without the predictive control, inverse dynamics is used in \textbf{No MPC} to output the control command. As a result, this controller usually traps in a local minimum to pursue the final goal while ignoring the long-term impacts.
Without the action prior, \textbf{No Prior} generates random action sequences within the huge space $\mathcal{A}$. This setup prevents the algorithm from finding a solution that satisfies the requirements in a finite iteration period.
Removing the cost function in \textbf{No Cost}, the solution from the optimization usually violates the constraints, which results in stopping an imitation episode in advance.

\begin{figure}
\centering
\centerline{\includegraphics[width=0.9\columnwidth]{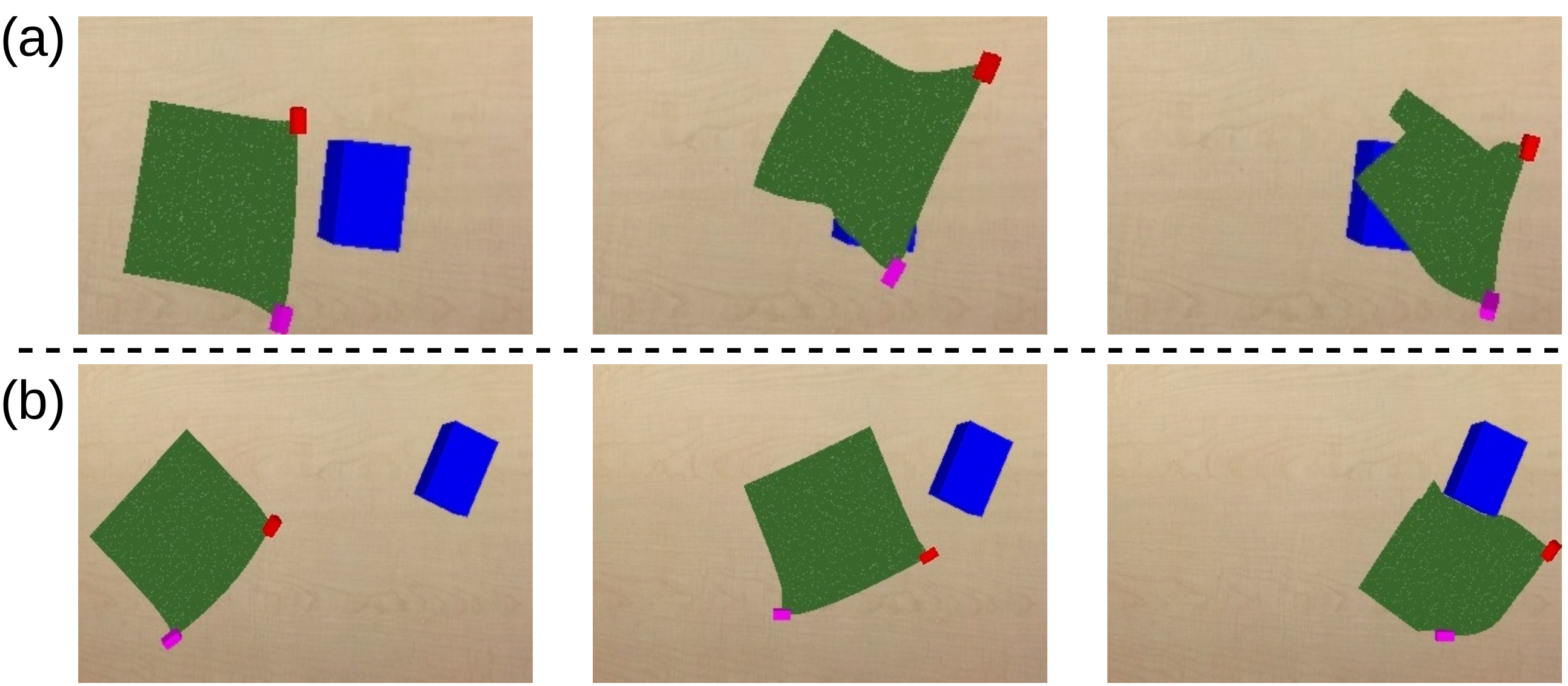}}
    \caption{Typical failure cases in simulation. (a) The fabric is manipulated to pass the rigid object. (b) The fabric reaches a state where it is difficult for the controller to acquire a feasible action.
    }
    \label{9_fail_simulation}
\end{figure}

\begin{figure}
\centering
\centerline{\includegraphics[width=\columnwidth]{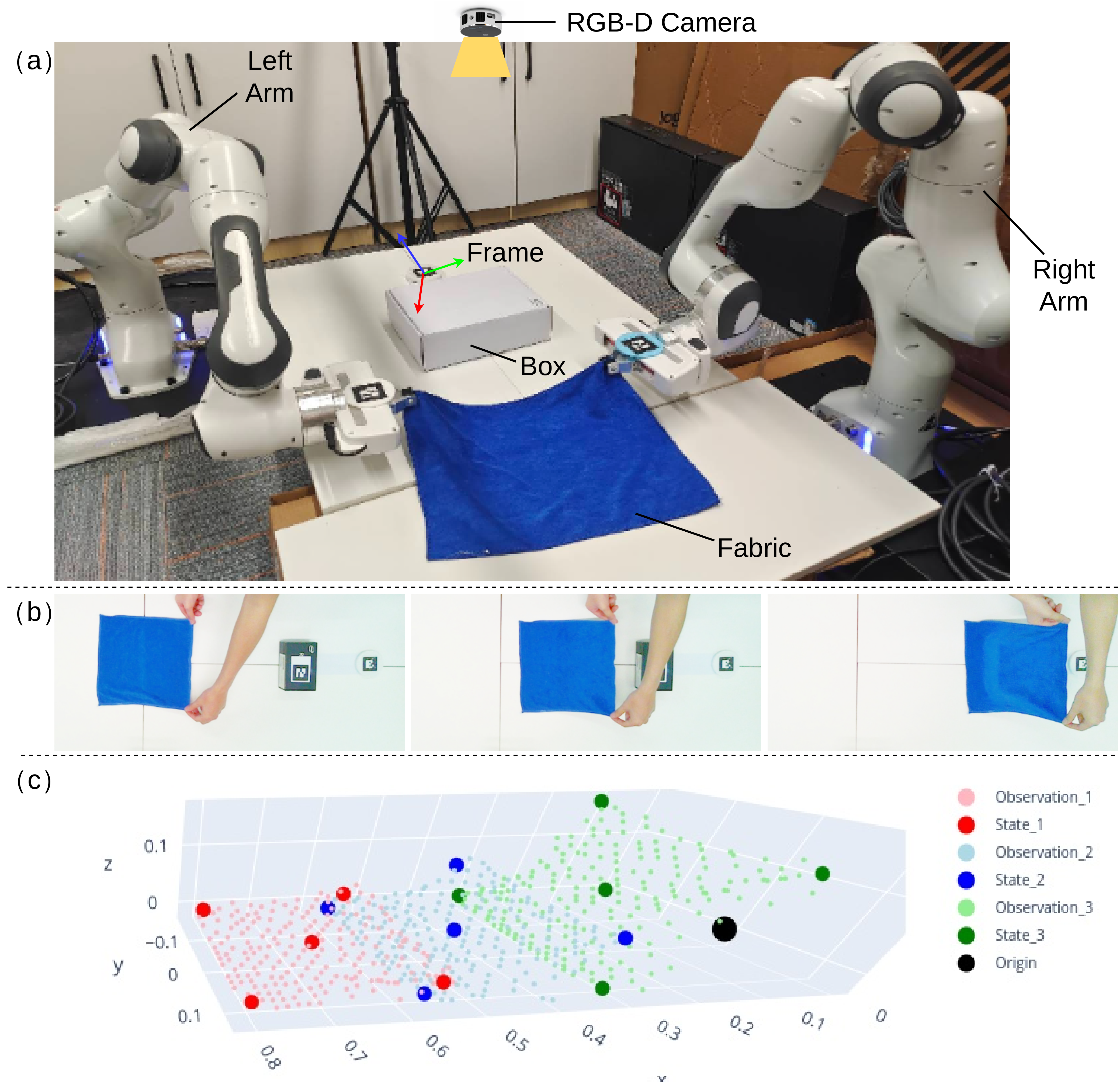}}
    \caption{Physical Scenarios to validate our approach. (a) Experimental Setup. (b) The process of the demonstration video provision with human hands. (c) Graphical illustration of the sequential demonstration.  
    }
    \label{10_setup_real}
\end{figure}

Although our approach is capable of handling the majority
of the challenging tasks, there are some situations when it fails. Fig. \ref{9_fail_simulation} presents two typical failure examples. The failure case in Fig. \ref{9_fail_simulation}(a) is mainly caused by the state alignment settings. 
The controller occasionally struggles in complex interactions with the rigid object since they are not considered in the dynamics models of priors. As a result, agents can not reach a state that satisfies the termination classifier and have to iteratively generate motions until passing the final state in the demonstration regardless of the penalization term in the reward function.
The failure in Fig. \ref{9_fail_simulation}(b) is mainly caused by the computation efficiency of the sampling-based MPC method.
To achieve real-time control, we define a finite searching period for each time-step, and our approach occasionally fails to find out an appropriate solution within this period. This phenomenon is more obvious when the fabric is close to the rigid object since the action space is more restricted.

\subsection{Physical Experiments}


\begin{table}
\centering
\caption{Results in physical experiments}
\begin{center}
\begin{tabular}{|c|c|c|cc|cc|} 
\hline
\textbf{Task} & Fabric & Rigidity & \multicolumn{2}{|c|}{Success Rate \%} \\
& (cm) & (cm) & Translation & Transformation \\ 
\hline
\textbf{Box-Small} & 30$\times$30 & 14$\times$9$\times$6 & 12/15 & 7/10 \\
\textbf{Hang} & 30$\times$30 & 17$\times$1$\times$20 & 12/15 & 8/10 \\
\textbf{Box-Big} & 30$\times$40 & 30$\times$15$\times$5 & 14/15 & 9/10 \\
\hline
\end{tabular}
\end{center}
\label{table_real}
\end{table}

\begin{figure*}
\centering
\centerline{\includegraphics[width=0.9\textwidth]{"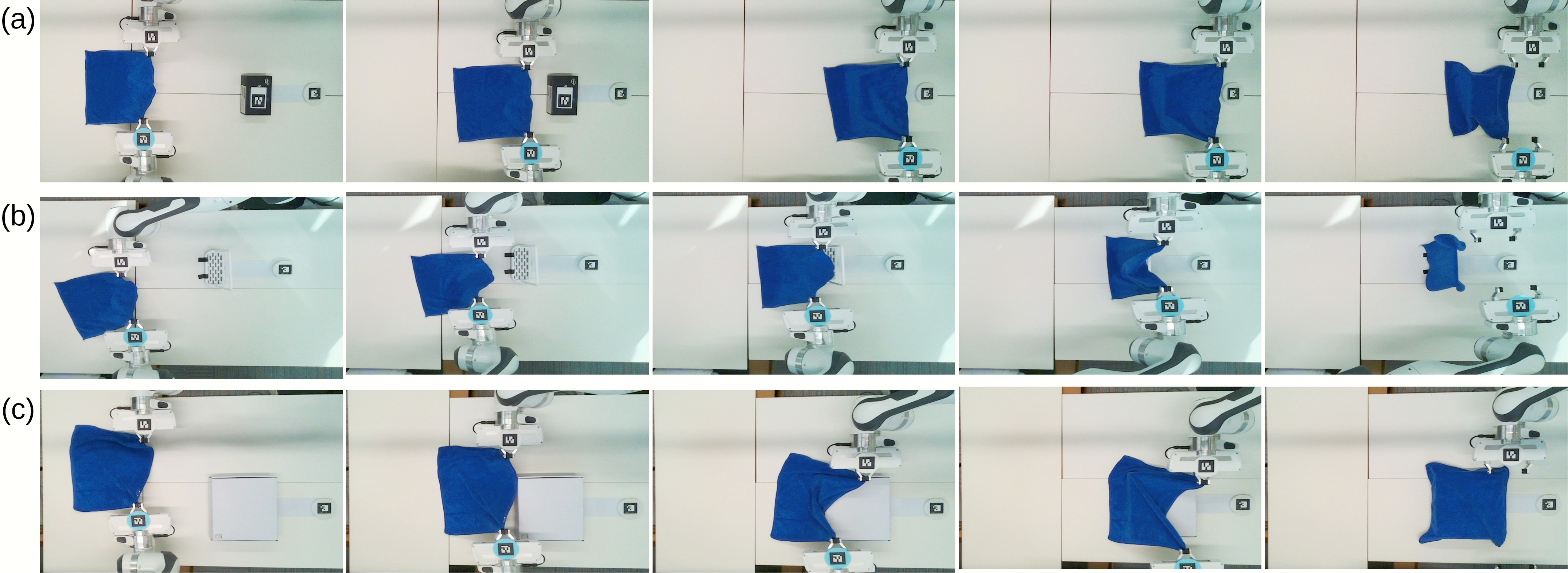"}}
    \caption{
    The scenarios of various contact-rich fabric manipulation tasks in real experiments. (a) \textbf{Box-Small}. (b) \textbf{Hang}. (c) \textbf{Box-Big}. 
    }
    \label{11_scene_real}
\end{figure*}

\begin{figure}
\centering
\centerline{\includegraphics[width=\columnwidth]{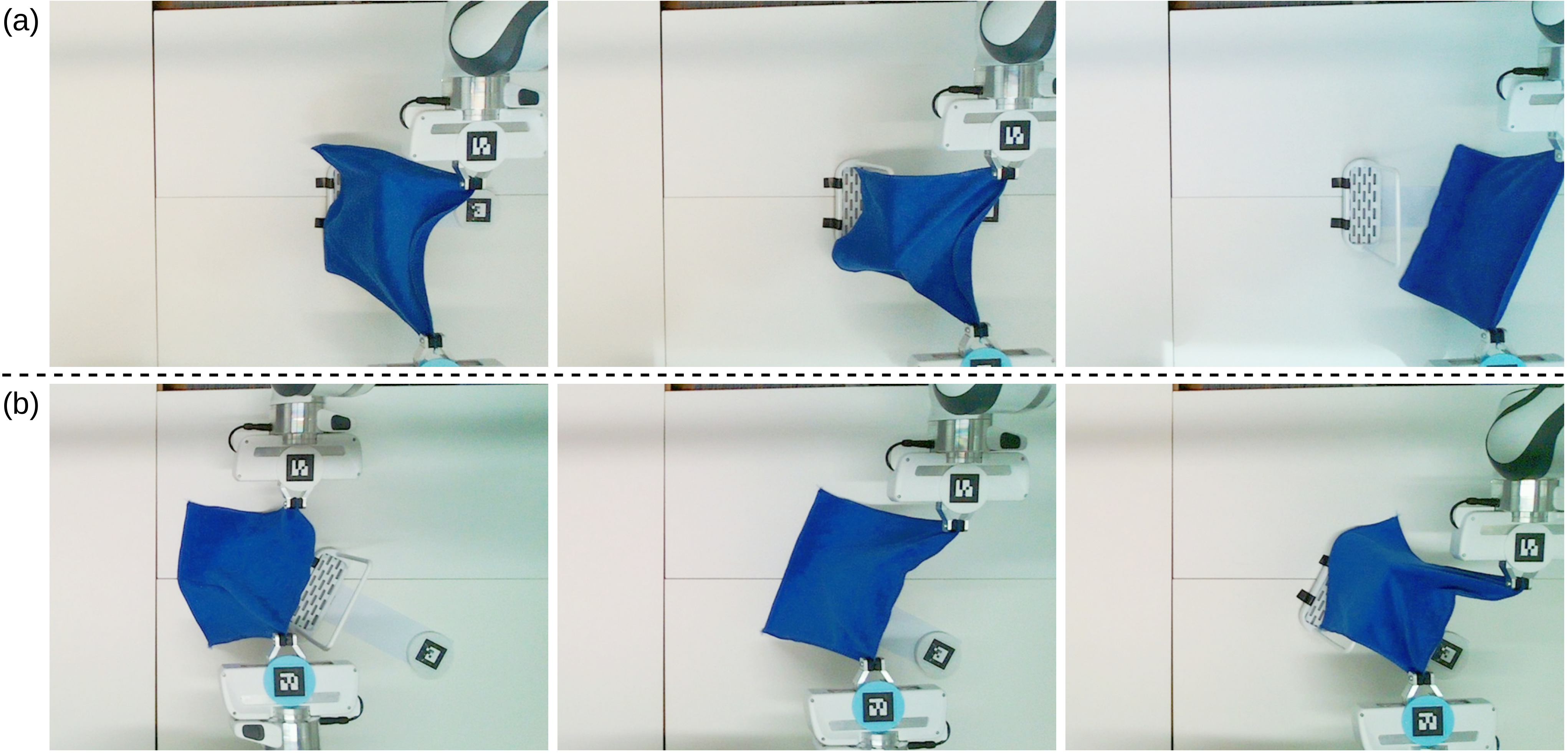}}
    \caption{Typical failure cases in real-world experiments. (a) The fabric is manipulated to pass the rigid object. (b) The fabric reaches a state where it is difficult for the state representation model $f_R(S^{[t]}|O^{[t]})$ to detect the customized keypoints.
    }
    \label{12_real_fail}
\end{figure}

This section gives the results of our proposed approach on real-world experiments. Fig. \ref{10_setup_real}(a) shows our robotic setup, which contains two Franka Emika Panda arms equipped with 2-fingered grippers and a visual feedback device. The RGB-D camera is fixed to provide a top-down perspective of the manipulation space. For a novel task, an operator provides a demonstration video about the bimanual manipulation with the hands, as shown in Fig. \ref{10_setup_real}(b). Note that we only collect the raw observations from the RGB-D camera 
and recognize the pose of the rigid object with its attached marker.
Then, we process the collected observations with our representation model $f_R(S^{[t]}|O^{[t]})$ to acquire the corresponding states, as shown in Fig. \ref{10_setup_real}(c).

To validate the effectiveness and robustness of our approach, we perform three kinds of contact-rich fabric manipulation tasks under different configurations. The details of various tasks are introduced in Table. \ref{table_real} and the individual successful scenarios are shown in Fig. \ref{11_scene_real}. 
Specifically, we consider three kinds of tasks, including \textbf{Box-Small}, \textbf{Hang}, and \textbf{Box-Big}. The major difference between \textbf{Box-Small} and \textbf{Box-Big} is the size of the fabric, which is used to evaluate the robustness of our algorithm in terms of the deformable object.
Without any additional explorations, our dual-arm robot is able to execute the imitation for a novel task immediately with our efficient skill-learning approach.
For each episode, dual arms start from a pre-grasp configuration and move the end-effectors when the control command is received. An episode is considered successful if the fabric establishes a corresponding symbolic relationship with the rigid object stably after releasing the grippers.

For each task, two levels of difficulty are analyzed: (1) \textbf{Translation}: Only change the position of the rigid object w.r.t. the situation in the demonstration. (2) \textbf{Transformation}: Change the pose of the rigid object arbitrarily within the workspace. 
The success rate for each individual contact-rich manipulation task is summarized in Table. \ref{table_real}. Each episode lasts approximately $20s$ to $40s$ in real-time, depending on the spatial distance between the initial configuration and the ultimate goal. These quantitative results illustrate the superior performance of our proposed approach in terms of efficiency and robustness. 

Fig. \ref{12_real_fail}(a) shows a failure case in physical experiments, which is similar to the scenario of Fig. \ref{9_fail_simulation}(a) in simulation. 
Since the exploration data is only collected in simulation, the transfer performance in reality is unavoidably affected due to their gap. 
Compared with the simulation results in Sec. IV-B, we acquire two distinct observations when analyzing the performance of the physical implementations.
The first one is that the success rate is decreased when the rotation transform is included. This is mainly because the kinematic feasibility of the robotic arms limits the manipulation space and brings additional difficulties for robots to align with the expert demonstrations. Considering the manipulability of robot arms in simulation instead of simplifying them as virtual anchors is beneficial to alleviate this problem.
The second one is that the occlusion phenomenon is unavoidable in robotic manipulations under grasping configuration, resulting in a lower keypoint detection accuracy. In the future, we consider leveraging temporal information to track the manipulated object during the entire manipulation.



\section{Conclusion}
In this paper, we propose an efficient robot skill learning approach through imitation from a single demonstration video. We learn general prior knowledge in one scene and a state alignment-based reward function for a new scene based on the provided demonstration video.
Robots are able to deploy the new scenario immediately by incorporating the above issues into our sampling-based MPC method. 
Since we do not assume the exact dynamic consistency between training and evaluation, the data collection procedure can be implemented in simulation to avoid time-consuming and high-cost procedures in real scenarios. 
Furthermore, the need for random physical explorations of robots is obviated, thereby mitigating potential hazards in unstructured environments.
The performance of the proposed approach is evaluated in the context of contact-rich fabric manipulation, in which two robotic arms need to cooperate in a constraint-aware manner.  
The results in both the simulation and real-world experiments show that robots are able to imitate the demonstration videos efficiently and safely.

In future work, we seek to leverage the temporal experience in the deployment of a task to improve the policy incrementally. In addition, integrating force sensing is beneficial to achieve a flexible and compliant control strategy for challenging robotic tasks.


\bibliographystyle{ieeetr}
\bibliography{ref}


 




\begin{IEEEbiography}
[{\includegraphics[width=1in,height=1.25in,clip,keepaspectratio]{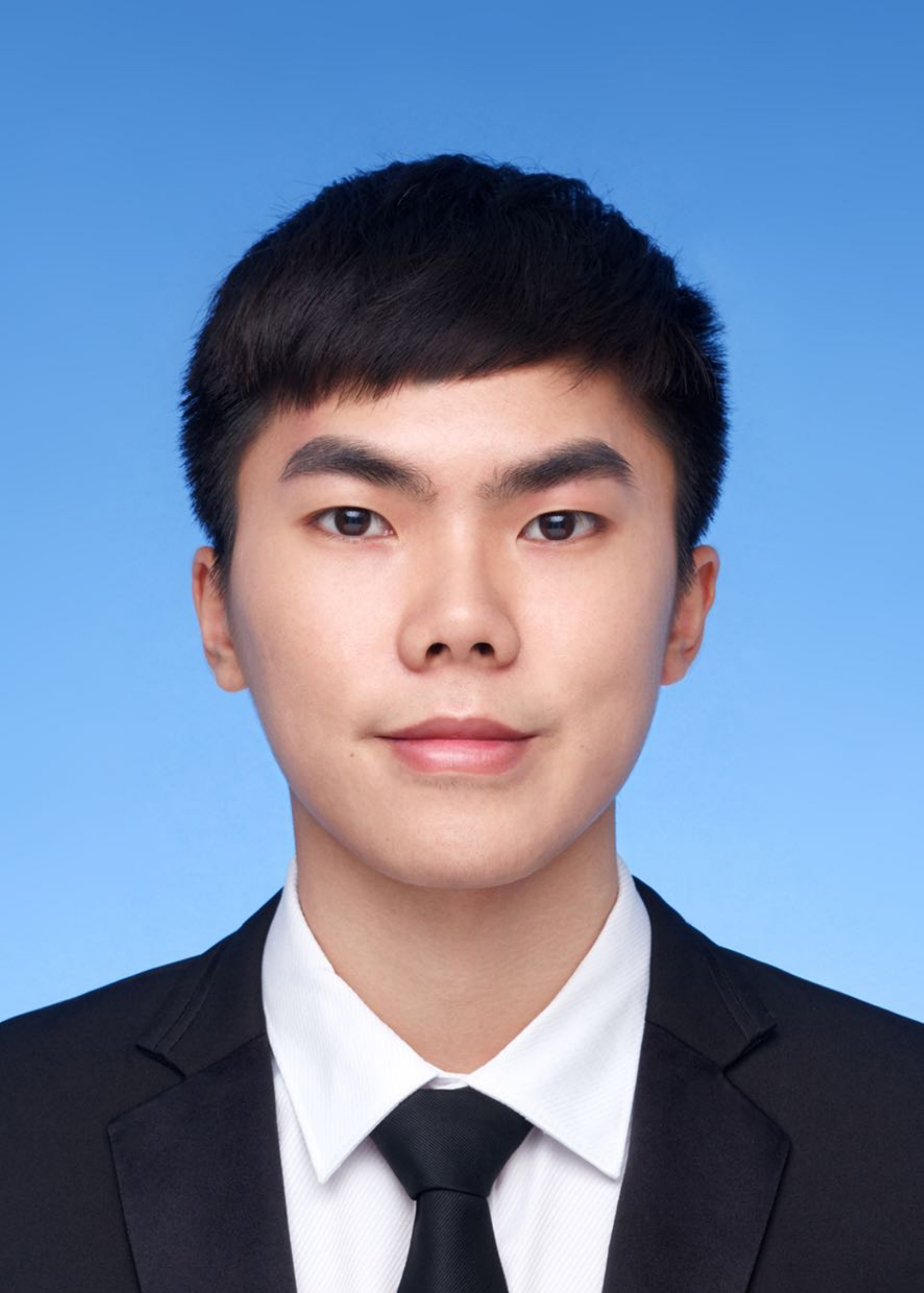}}]
{Shengzeng Huo} received the B.S. degree in vehicle engineering from the South China University of Technology, Guangzhou, China, in 2019. He is currently pursuing the Ph.D. degree in the Department of Mechanical Engineering from The Hong Kong Polytechnic University, Hong Kong. His research interests include bimanual manipulation, deformable object manipulation, and robot learning.  
\end{IEEEbiography}

\begin{IEEEbiography} [{\includegraphics[width=1in,height=1.25in,clip,keepaspectratio]{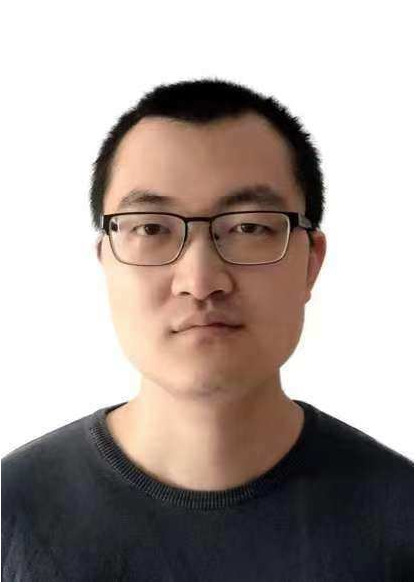}}]
	{Anqing Duan} received his Ph.D. degree in robotics from the Italian Institute of Technology and the University of Genoa in 2021.
	He is currently Research Associate with the Robotics and Machine Intelligence Laboratory at The Hong Kong Polytechnic University. His research interests include robotics and learning-based control.
\end{IEEEbiography}

\begin{IEEEbiography} [{\includegraphics[width=1in,height=1.25in,clip,keepaspectratio]{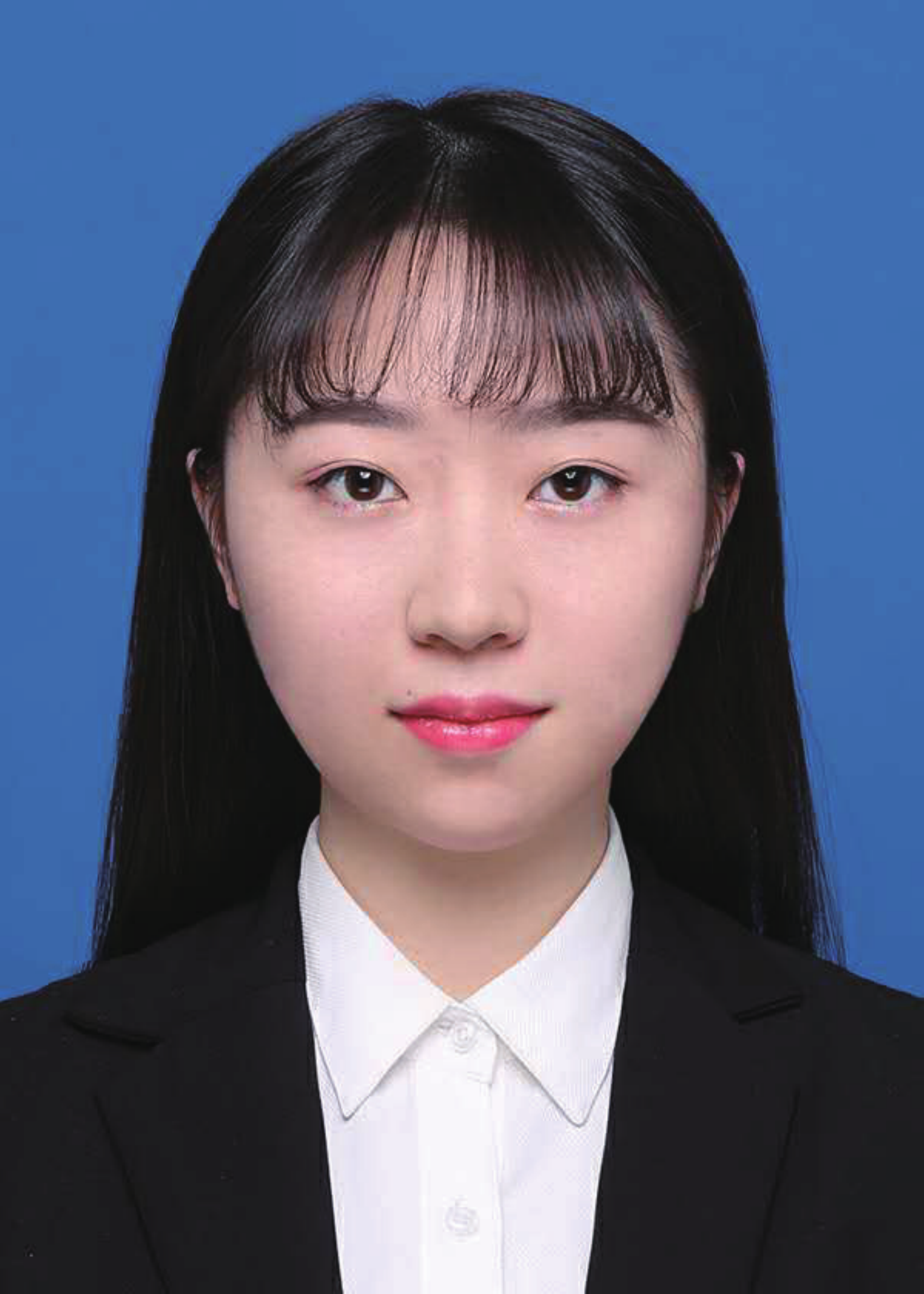}}]
{Lijun Han} received the B.S. degree in automation from Tianjin University, Tianjin, China, in 2019. She is currently pursuing the Ph.D. degree in control technology and control engineering with the Department of Automation, Shanghai Jiao Tong University, Shanghai, China. 
 
Her research interests include visual servoing, robot control, and surgical robots.
\end{IEEEbiography}


\begin{IEEEbiography}
[{\includegraphics[width=1in,height=1.25in,clip,keepaspectratio]{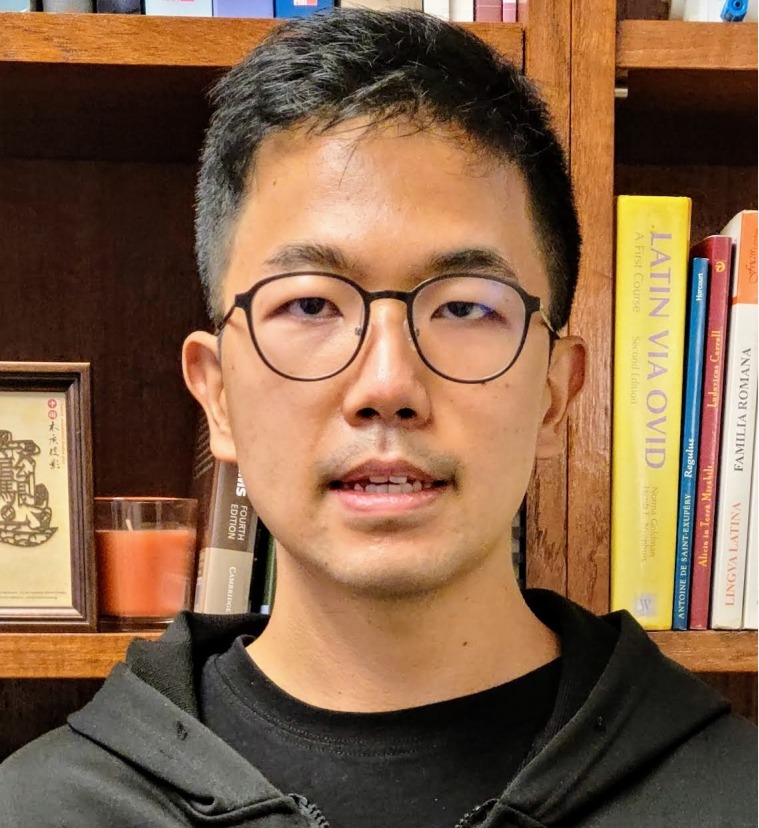}}]
{Luyin Hu} received his BEng. and MPhil. degree in Mechanical Engineering from The Hong Kong Polytechnic University, KLN, Hong Kong in 2019 and 2022. He is currently a research assistant at the Centre for Transformative Garment Production, Hong Kong. His research interests include servomechanisms, control system design, and physics simulation. 
\end{IEEEbiography}


\begin{IEEEbiography}
[{\includegraphics[width=1in,height=1.25in,clip,keepaspectratio]{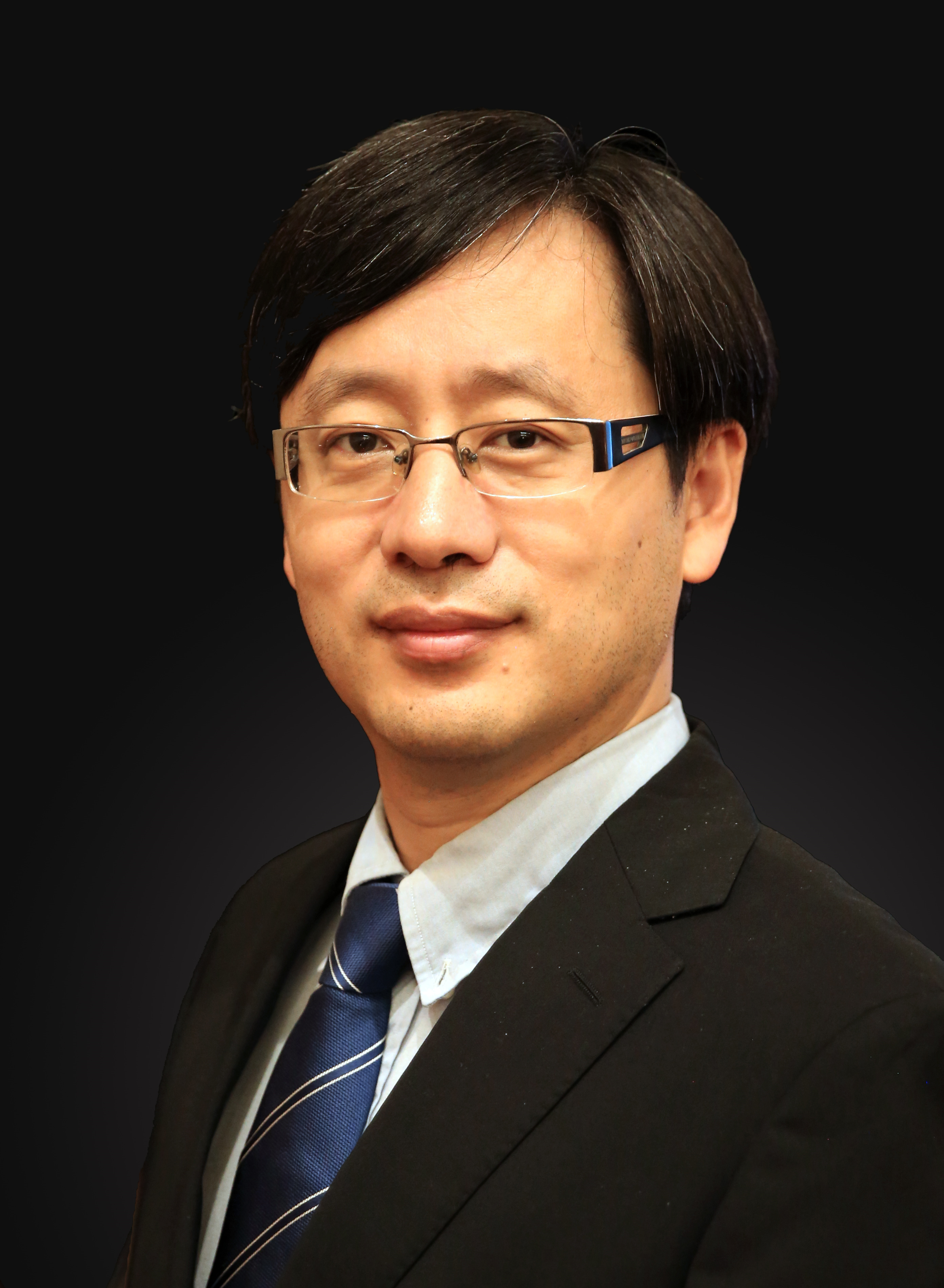}}]
{Hesheng Wang} (Senior Member, IEEE) received the M.Phil.
and Ph.D. degrees in automation and computer-aided engineering from The Chinese University of Hong Kong, Hong Kong, in 2004 and 2007, respectively.
He is currently a Professor with the Department of
Automation, Shanghai Jiao Tong University, Shanghai, China. His current research interests include visual servoing, service robot, computer vision, and autonomous driving. Dr. Wang is an Associate Editor of IEEE Transactions on Automation Science and Engineering, IEEE Robotics and Automation Letters, Robotic Intelligence and Automation and the International Journal of Humanoid Robotics, a Technical Editor of the IEEE/ASME Transactions on Mechatronics, an Editor of Conference Editorial Board (CEB) of IEEE Robotics and Automation Society. He served as an Associate Editor of the IEEE Transactions on Robotics from 2015 to 2019. He was the General Chair of IEEE ROBIO 2022 and IEEE RCAR 2016, and the Program Chair of the IEEE ROBIO 2014 and IEEE/ASME AIM 2019. He will be the General Chair of IEEE/RSJ IROS 2025.
\end{IEEEbiography}

\begin{IEEEbiography}
[{\includegraphics[width=1in,height=1.25in,clip,keepaspectratio]{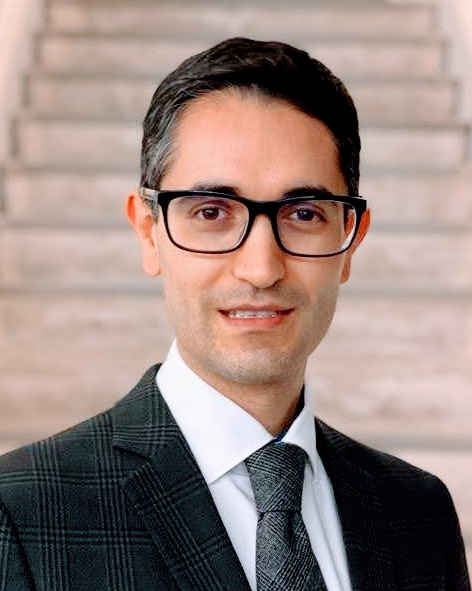}}] {David Navarro-Alarcon} (Senior Member, IEEE) received the Ph.D. degree in mechanical and automation engineering from The Chinese University of Hong Kong, NT, Hong Kong, in 2014. 
Since 2017, he has been with The Hong Kong Polytechnic University, where he is currently an Assistant Professor with the Department of Mechanical Engineering. His research interests include robotics, control, and nonlinear dynamics. He currently serves as an Associate Editor of the \textsc{IEEE Transactions on Robotics}.
\end{IEEEbiography}

\end{document}